\documentclass{article}

    \usepackage[nonatbib,preprint]{neurips_2020}

\usepackage[utf8]{inputenc} 
\usepackage[T1]{fontenc}    
\usepackage{hyperref}       
\usepackage{multirow}
\usepackage{subcaption}
\usepackage[export]{adjustbox}
\usepackage{bm}
\usepackage{wrapfig}
\usepackage{wrapft}
\usepackage{ntheorem}
\usepackage{algorithmicx,algorithm}
\usepackage{algpseudocode}
\usepackage{amsmath}
\newtheorem{theorem}{Theorem}[subsection]
\usepackage{url}            
\usepackage{booktabs}       
\usepackage{amsfonts}       
\usepackage{nicefrac}       
\usepackage{microtype}      
\usepackage{booktabs} \usepackage{amssymb}
\renewcommand{\arraystretch}{1.3}
\usepackage{graphicx}
\usepackage{paralist}
\title{Uniform Interpolation Constrained Geodesic Learning on Data Manifold}

\author{
  Cong Geng, Jia Wang, Li Chen,
   Wenbo Bao, Chu Chu, Zhiyong Gao\\
  Institute of Image Communication and Network Engineering\\
   Shanghai Jiao Tong University, Shanghai, China\\
  \texttt{\{gengcong,jiawang,hilichen,baowenbo,chu\_chu,zhiyong.gao\}@sjtu.edu.cn} 
}

\begin{document}

\maketitle

\begin{abstract}
  In this paper, we propose a method to learn a minimizing geodesic within a data manifold. 
  Along the learned geodesic, our method is able to generate high-quality interpolations between two given data samples.
  Specifically, we use an autoencoder network to map data samples into the latent space and perform interpolation via an interpolation network.
  We add prior geometric information to regularize our autoencoder for the convexity of representations so that for any given interpolation approach, the generated interpolations remain within the distribution of the data manifold. The Riemannian metric on data manifold is induced by the canonical metric in the Euclidean space in which the data manifold is isometrically immersed.
  Based on this defined Riemannian metric, we introduce a constant-speed loss and a minimizing geodesic loss to regularize the interpolation network to generate uniform interpolations along the learned geodesic on the manifold. 
  We provide a theoretical analysis of our model and use image interpolation as an example to demonstrate the effectiveness of our method.
\end{abstract}
\section{Introduction}
With the success of deep learning on approximating complex non-linear functions,
representing data manifolds using neural networks has received considerable interest and been studied
from multiple perspectives. Autoencoders (AE)~\cite{kingma2014AutoEncoding}, generative adversarial network~(GAN)~\cite{goodfellow2014generative} and their combinations~\cite{makhzani2015adversarial,tolstikhin2018Wasserstein} are notable generative  models for learning the geometry structure or the distribution of data on the manifold. However, they usually fail to generate a convex latent representation when confronting with even simple curved manifolds such as swiss-roll and S-curve. For geodesic learning, Arvanitidis et al.~\cite{arvanitidis2018Latent,yang2018Geodesic} and Chen et al.~\cite{chen2018Metrics,chen2019fast} both present a magnification factor~\cite{bishop1997magnification} to help find the shortest path that follows the regions of high data density in the latent space as the learned geodesic. But these methods explore geodesics on latent space rather than on data space and  lack strict guarantees for
geodesics in mathematics.

In this paper, we propose a method to capture the geometric structure of a data manifold and to find smooth geodesics between data samples on the manifold. First, motivated by DIMAL~\cite{pai2019DIMAL}, we introduce a framework combining an autoencoder with traditional manifold learning methods to explore the geometric structure of a data manifold. In contrast to existing combinations of autoencoders and GANs, we resort to the classical manifold learning algorithms to obtain an approximation of geodesic distance.
Then the encoder and decoder are optimized with this approximation being the constraint. 
Using this method, we alleviate the problem that the latent distribution is non-convex on curved manifolds,  
which means our encoder can unfold the data manifold to flattened latent representations.

Second, we propose a geodesic learning method to establish smooth geodesics between data samples on a data manifold. The Riemannian metric on the data manifold can be induced by the canonical metric in the Euclidean space in which the manifold is isometrically immersed. 
Existing methods cannot guarantee a uniform interpolation along the geodesic. 
We parameterize the generated curve by a parameter $t$ and propose a constant-speed loss to achieve a uniform interpolation by making $t\in [0,1]$ be an arc-length parameter of the interpolation curve. 
According to Riemannian geometry~\cite{carmo1992riemannian}, we propose a geodesic loss to force the interpolation network to generate points along a geodesic. Another key factor is a minimizing loss to make the geodesic be the minimizing one. Our contributions are summarized as follows:
\begin{compactitem}
	
	\item We propose a framework in which an autoencoder and an interpolation network are introduced to explore the manifold structure. The autoencoder network promotes convex latent representations achieved by adding some prior geometric information to it, so that the generated samples are able to remain within the distribution of the data manifold.
	\item We propose a constant-speed loss and a minimizing geodesic loss for the interpolation network to generate the minimizing geodesic on the manifold given two endpoints. We parameterize each geodesic by an arc-length parameter $t$, such that we can generate points moving along the minimizing geodesic with constant speed and thus fulfill a uniform interpolation.
\end{compactitem}
\section{Related Work}

Traditionally, manifold learning methods aim to infer latent representations that can capture the intrinsic geometric structure of data. 
They are generally applied in dimensionality reduction and data visualization. 
Classical nonlinear manifold learning algorithms such as Isomap~\cite{isomap}, locally linear embedding~\cite{LLE} and local tangent space alignment~(LTSA)~\cite{ltsa} preserve local structures in small neighborhoods and derive the intrinsic features of data manifolds.

The interpolation between data samples has attracted wide attention because it can achieve a smooth transition from one sample to another, such as intermediate face generation and path planning.
There are several works of literature focusing on the interpolation of data manifold.
Agustsson et al.~\cite{agustsson2019Optimal} propose to use distribution matching transport maps to obtain analogous latent space interpolation which preserves the prior distribution of the latent space, while minimally changing the original operation. 
It can generate interpolated samples with higher quality but it does not have an encoder to map the given data samples to the latent space. Several other works~\cite{berthelot2019Understandinga,sainburg2018generative,chen2019homomorphic} that combine autoencoders with GANs are proposed to generate interpolated points on data manifold. But limited by the ability of GAN, they usually fail to generate a convex latent representation when confronting with curved manifolds such as swiss-roll and S-curve. To avoid this problem, Arvanitidis et al.~\cite{arvanitidis2018Latent,yang2018Geodesic} and Chen et al.~\cite{chen2018Metrics} both present that the magnification factor~\cite{bishop1997magnification} can be seen as a measure of the local distortion of the latent space. This helps the generated geodesics to follow the regions of high data density in the latent space.  
Chen et al.~\cite{chen2019fast} also find the shortest path in a finite graph of samples as the geodesics on the data manifold. This finite  graph is built in the latent space using a binary tree data structure with edge weights based on Riemannian distances.

\section{Preliminaries}
\label{2.1}

We briefly present the essentials of Riemannian geometry. For a thorough presentation, see~\cite{carmo1992riemannian}. Suppose $M$ is a Riemannian differentiable manifold. $\alpha:(-\epsilon,\epsilon) \to M$ is a differentiable curve in $M$. Suppose that $\alpha(0)=p \in M$, the tangent vector to the curve $\alpha$ at $t=0$ is a function $\alpha'(0)$ defined on the arbitrary function $f$ on $M$ given by $\alpha'(0)f=\frac{d(f\circ\alpha)}{dt}\bigg|_{t=0}$.
The set of all tangent vectors to $M$ at $p$ will be indicated by tangent space $T_pM$. Let $(U,x)$ be a parametrization of $M$ at $p$ and for $\forall q \in U$, suppose $x^{-1}(q)= (z^1(q), z^2(q),\cdots,z^n(q))$ is the local coordinate of $q$. Then the tangent vector at $p$ of the coordinate curve $\frac{\partial }{\partial z^i}\big|_{p}$ is defined as$\frac{\partial }{\partial z^i}\big|_{p}(f)=\frac{\partial (f\circ x)}{\partial z^i}\big|_p$. $\{\frac{\partial }{\partial z^i}\arrowvert_p, 1\leqslant i\leqslant n\}$ spans a tangent space $T_pM$ at $p$. If the tangent vector is orthonormal with each other, $\{\frac{\partial }{\partial z^i}\arrowvert_{p}, 1\leqslant i\leqslant n\}$ is the orthonormal basis of tangent space.

Let $M^m$ and $N^n$ be two differentiable manifolds. Given a differentiable mapping $\varphi:M \to N$, we can define a differential of $\varphi$ at $p$ as linear mapping $d\varphi_p:T_pM \to T_{\varphi(p)}N$ given by $d\varphi(p)(v)(f)=v(f\circ\varphi)$, for $\forall f \in C_{\varphi(p)}^\infty$. If $d\varphi(p): T_pM \to T_{\varphi(p)}N$ is injective for all $p \in M$, then the mapping $\varphi$ is said to be an immersion.

A Riemannian manifold is a differentiable manifold equipped with a given Riemannian metric which assigns on each point $p$ of $M$ an inner product $\langle ,\rangle_p$ (i.e. a symmetric, bilinear and positive-definite form) on the tangent space $T_pM$. Furthermore, an isometric immersion $\varphi:M \to N$ is an immersion satisfying $\big\langle u,v \big\rangle_p= \big\langle df_p(u),df_p(v) \big\rangle_{f(p)}$ for $\forall p \in M$ and $\forall u,v \in T_pM$. 

A geodesic is a parametrized curve $\gamma: I \to M$ satisfying $\frac{D}{dt}(\frac{d\gamma}{dt})=0$ for all $t \in I$. $\frac{D}{dt}$ is the covariant derivative of a vector field along $\gamma$ \cite{carmo1992riemannian}. $\frac{d\gamma}{dt}$ is the tangent vector at $t$.

\section{Manifold Reconstruction}

\begin{figure}[!t]
	\footnotesize
	\centering
	\renewcommand{\tabcolsep}{1pt} \renewcommand{\arraystretch}{0.9} \begin{tabular}{cccccc}
		\includegraphics[width=0.16\linewidth]{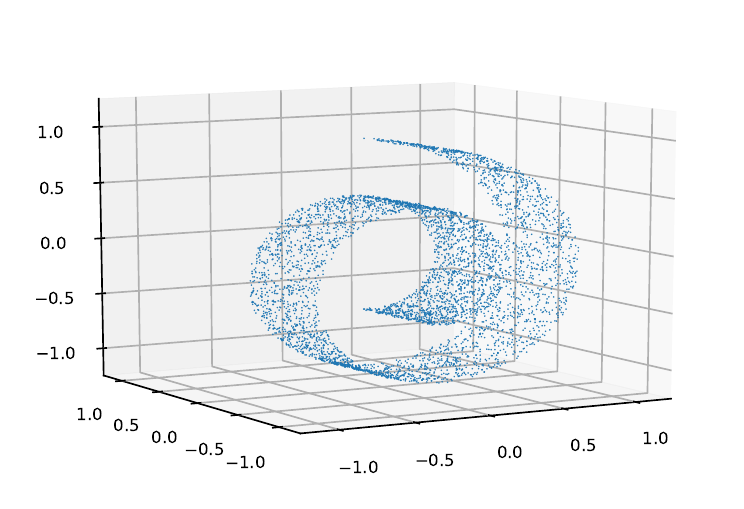} &
		\includegraphics[width=0.16\linewidth]{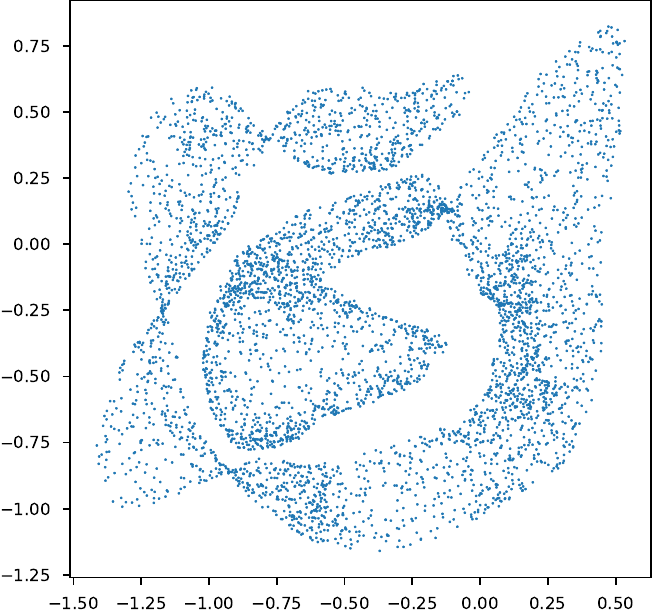} &
		\includegraphics[width=0.16\linewidth]{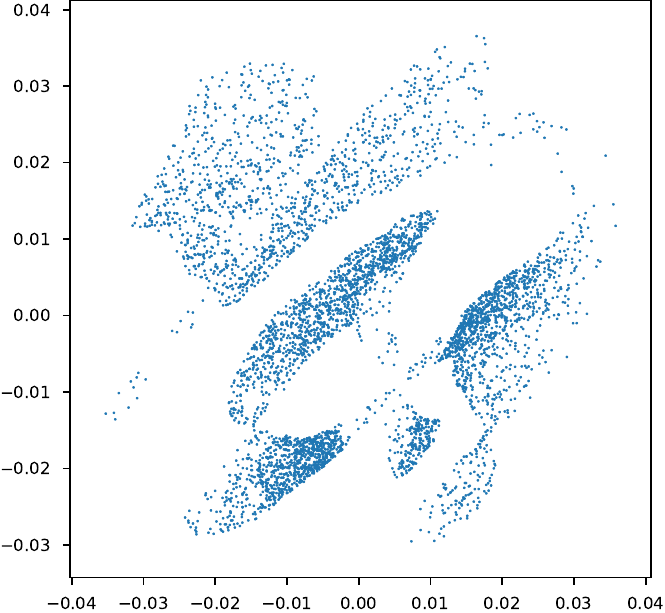}&
		\includegraphics[width=0.15\linewidth]{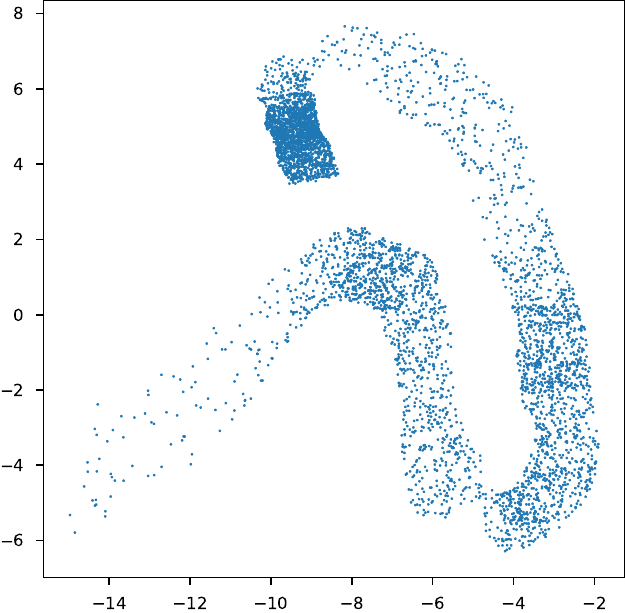}&
		\includegraphics[width=0.15\linewidth]{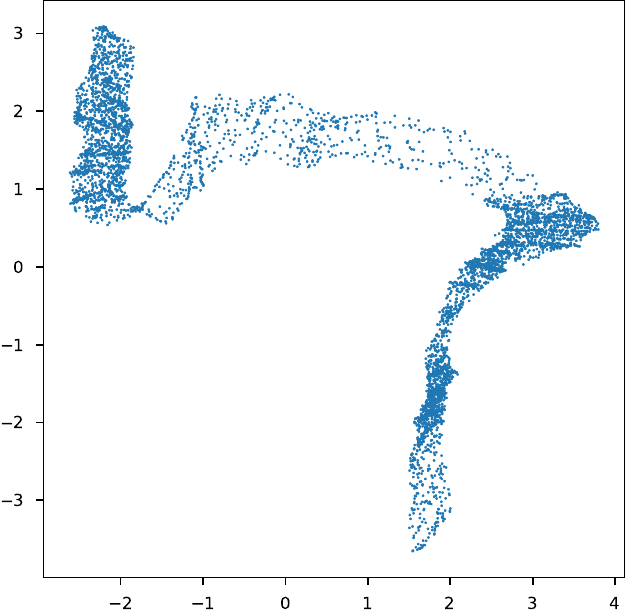}&
		\includegraphics[width=0.16\linewidth]{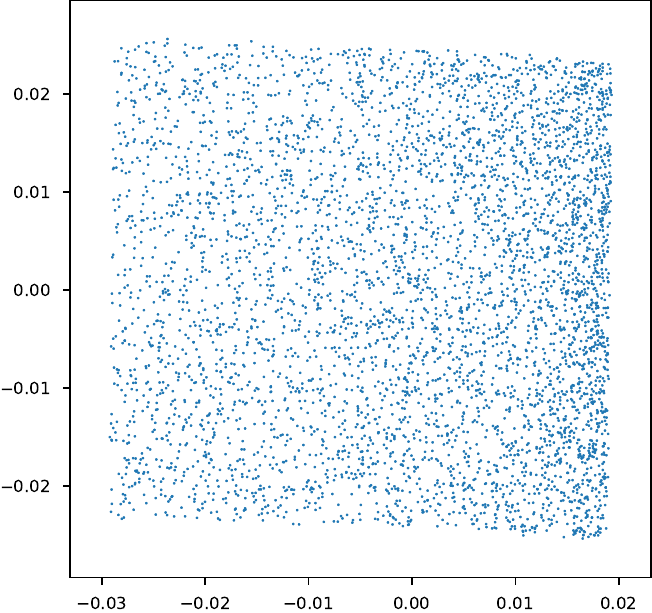}
		\\
		(a) swiss-roll & (b) AAE~\cite{makhzani2015adversarial} & (c) GAIA~\cite{sainburg2018generative}&(d) ACAI~\cite{berthelot2019Understandinga}& (e) Chen~\cite{chen2019homomorphic} & (f) ours\\
	\end{tabular}
	\vspace{-5pt}
	\caption{The encoding results of different methods for swiss-roll.
	}
	\label{encodeing_results} \vspace{-10pt}
\end{figure}

Original AE and some combinations between AE and GANs can generate high-quality samples from specific latent distributions on high-dimensional data manifold. But they fail to get convex  embeddings on some curved surfaces with changing curvature such as swiss-roll or S-curve. Some other works~\cite{berthelot2019Understandinga,sainburg2018generative,chen2019homomorphic} were proposed to generate interpolations within the distribution of real
data by distinguishing interpolations from real samples. But they generate unsatisfying samples on the above-mentioned low-dimensional manifolds. The encoding results can be seen in Fig.~\ref{encodeing_results}. The reason for this problem may be the insufficient ability of GANs and the decoder of AE. The discriminator of GANs can only distinguish the similarities of distributions between generated samples and real ones and the autoencoders are prone to put the curvature information of data manifold in latent representations to make the network simple as possible. So the latent embeddings also have a changing curvature to save the curvature information. It may result in the non-convex representations.


In our method, we add some prior geometric information obtained by traditional manifold learning approaches to encoding a convex latent representation. Traditional nonlinear manifold learning approaches such as Isomap~\cite{isomap}, LLE~\cite{LLE} and LTSA~\cite{ltsa} are classical algorithms to get a convex embedding by unfolding some curved surfaces. We apply them to our method by adding regularization to the autoencoder. 
The loss function of the autoencoder can be written as:
\begin{equation}
\label{1}
L_{AE}=\big\|D\big(E(x)\big)-x\big\|+\frac{1}{M^2}\sum_{i,j}\Big(\big\|E(x_i)-E(x_j)\big\|_2-DKS_{ij}\Big)^2,
\end{equation}
where $x$ is the input sampled on the data manifold. $M$ denotes the number of input $x$. $DKS_{ij}$ represents the approximated geodesic distance between $E(x_i)$ and $E(x_j)$. Encoder $E$ and Decoder $D$ of an autoencoder are trained to minimize the above loss $L_{AE}$. $DKS$ can be obtained by building a graph from the data with k-nearest neighbors for each node and computing the corresponding pairwise geodesic distances using traditional manifold learning algorithms. For more details we refer the interested reader to~\cite{isomap,LLE,ltsa}. With $DKS_{ij}$ as an expected approximation, the encoder is forced to train towards obtaining a convex latent representation while the decoder is forced to learn the lost curvature information on latent embeddings. Behaviors induced by the $L_{AE}$ loss can be observed in Fig.~\ref{encodeing_results}: the swiss-roll can be flattened on 2-dimensional latent space. For our experiments, we choose LTSA to compute the approximated geodesic distance for regularizing the autoencoder because of its learned local geometry which views the neighborhood of a data point as a tangent space to flat the manifold.
\section{Geodesic Learning}
\label{headings}
\vspace{-10pt}
In our model, we denote $\mathcal{X}$ as a data manifold. The Riemannian metric on $\mathcal{X}$ can be induced from the canonical metric on the Euclidean space $\mathbb{R}^N$ to guarantee the immersion is an isometric immersion. Thus to obtain a geodesic on manifold $\mathcal{X}$, we can use the Riemannian geometry on $\mathbb{R}^N$ and the characteristics of isometric immersion. 
\subsection{Interpolation Network}
We produce geodesics on manifold $\mathcal{X}$ by interpolating in the latent space and decoding them into data space. The method of interpolating in the latent space can be varied in different situations. The simplest interpolation is linear interpolation as $z=(1-t)\cdot z_1+t\cdot z_2$. For geodesic learning, linear interpolation is not applicable in most situations. Yang et al.~\cite{yang2018Geodesic} propose to use the restricted class of quadratic functions and Chen et al.~\cite{chen2018Metrics} apply a neural network to parameterize the geodesic curves. For our interpolation network, we employ polynomial functions similar to Yang's approach. The difference is that we employ cubic functions to parameterize interpolants considering the diversity of latent representations, i.e., $c(t)=at^3+bt^2+ct+d$.
Therefore, a curve generated by our interpolation network has four $m$-dimensional free parametric vectors $a, b, c$ and $d$, where $m$ is the dimension of latent coordinates. In practice, we train a geodesic curve $c(t)$ that connects two pre-specified points $z0$ and $z1$, so the function should be constrained to satisfy $c(0) = z0$ and $c(1) = z1$. We initialize our interpolation network by setting $a=0$ and $b=0$ to make the initial interpolation be a linear interpolation and perform the optimization using gradient descent. More details can be referred to in Yang's paper~\cite{yang2018Geodesic}.

\subsection{Constant-Speed Loss}
\label{Constant Speed Loss}
We can produce interpolations along a curve on manifold $\mathcal{X}$ by decoding the output $c(t)$ of the interpolation network as $t$ from 0 to 1. We expect the parameter $t$ to be an arc-length parameter which means that the parameter $t$ is proportional to the arc length of the curve $\gamma(t)$. To realize this, the following theorem can provide theoretical support.
\begin{theorem}\label{4.1} Suppose $\mathcal{X} \subset \mathbb{R}^N$ is a Riemannian differentiable manifold. The Riemannain metric on $\mathcal{X}$ is induced from the canonical metric on $\mathbb{R}^N$. If $\gamma: I\to \mathcal{X}$ is a geodesic on $\mathcal{X}$,  $\big[x_1(t),x_2(t),\cdots,x_N(t)\big]$ is the Cartesian coordinate of $\gamma(t)$ in $\mathbb{R}^N$, then $\sqrt{\sum_{i}\big(\frac{dx_i(t)}{dt}\big)^2}$ is a constant, for$~\forall t\in I$.
\end{theorem}
As stated in Theorem~\ref{4.1}, the length of the tangent vector along a geodesic is constant. Let $G(t)=D(c(t))$ denote the output of the decoder taking the interpolation curve as input, we design the following constant-speed loss as:
\begin{equation}
L_{conspeed}=\frac{1}{n}\sum_{i=1}^{n}\Big(\frac{\|G'(t_i)\|_2}{mean_t\big(\|G'(t)\|_2\big)}-1\Big)^2,
\end{equation}
where $n$ denotes the number of sampling points and $mean_t\big(\|G'(t)\|_2\big)=\frac{1}{n} \sum\limits_{i=1}^n\big\|G'(t_i)\big\|_2$.
$G'(t)$ denotes the derivative of the output $G(t)$ with respect to $t$. It can be viewed as the velocity at $t$ along $G(t)$ and $\big\|G'(t)\big\|_2$ represents the magnitude of this velocity corresponding to $\sqrt{\sum_{i}\big(\frac{dx_i(t)}{dt}\big)^2}$ in Theorem~\ref{4.1}.  Therefore, from a certain angle, constant-speed loss makes $G(t)$ have a constant speed with $t$ moving from 0 to 1. 

\subsection{Minimizing Geodesic Loss}
\label{Minimizing Geodesic Loss}

After guaranteeing the output curve $G(t)$ of our decoder has a constant speed, we need to let the curve $G(t)$ be a geodesic. We have the following theorem to ensure a curve is a geodesic. 
\begin{theorem}\label{4.2} Suppose $\mathcal{X} \subset \mathbb{R}^N$ is a Riemannian differentiable manifold. The Riemannain metric on $\mathcal{X}$ is induced from the canonical metric on $\mathbb{R}^N$.  $\gamma: I\to \mathcal{X}$ is a curve on $\mathcal{X}$, $(U,h)$ is a system of coordinates of $\mathcal{X}$ with $\gamma(t)\subset h(U)$. $[z^1,z^2,\cdots,z^m]$ is the local coordinate of $h(U)$, $\big[x_1(t),x_2(t),\cdots,x_N(t)\big]$ is the Cartesian coordinate of $\gamma(t)$ in $\mathbb{R}^N$, then $\gamma(t)$ is a geodesic on $\mathcal{X}$, if and only if:
\begin{equation}
[x_1''(t),x_2''(t),\cdots,x_N''(t)]\cdot \Big[\frac{\partial h}{\partial z^1}\Big|_{\gamma(t)},\frac{\partial h}{\partial z^2}\Big|_{\gamma(t)},\cdots,\frac{\partial h}{\partial z^m}\Big|_{\gamma(t)}\Big]=\bm{0}, for~\forall t\in I.
\end{equation}
\end{theorem}

Theorem~\ref{4.2} demonstrates a curve is a geodesic if and only if its second derivative with respect to parameter $t$ is orthogonal to the tangent space. In practice, we can assume our encoder maps a point on the manifold $\mathcal{X}$ to its local coordinates. So based on Theorem~\ref{4.2}, we are able to optimize the following problem as the geodesic loss:
\begin{equation}
L_{geo}=\frac{1}{n}\sum_{i=1}^{n}\big\|G''(t_i)\cdot D'\big(c(t_i)\big)\big\|_2,
\end{equation}
where $D$ is the function of decoder and $D'\big(c(t_i)\big)$ denotes the derivative of $D$ at $c(t_i)$. $G''(t)$ is a $N$-dimensional vector corresponding to $[x_1''(t),x_2''(t),\cdots,x_N''(t)]$ and $D'\big(c(t)\big)$ is an $N\times m$  matrix corresponding to $\big[\frac{\partial h}{\partial z^1}\big|_{\gamma(t)},\frac{\partial h}{\partial z^2}\big|_{\gamma(t)},\cdots,\frac{\partial h}{\partial z^m}\big|_{\gamma(t)}\big]$ in Theorem~\ref{4.2}. Geodesic loss and constant-speed loss jointly force curve $G(t)$ to have zero acceleration. That is, $G(t)$ is a geodesic on data manifold. But geodesic connecting two points may not be unique, such as the geodesic on a sphere. According to Theorem~\ref{A1} in supplementary materials, the minimizing geodesic is the curve with minimal length connecting two points. Thus our model takes advantage of the minimal length to ensure $G(t)$ is a minimizing geodesic. We approximate the curve length using the summation of velocity at $t_i$. The minimizing loss is proposed to minimize curve length as:
\begin{equation}
L_{min}=\sum_{i=1}^{n}\big\|G'(t_i)\big\|_2.
\end{equation}

For implementation, we use the following difference approximation to reduce the computational burden:
\begin{equation}
\label{6}
G'(t)\approx \frac{G(t+\Delta t)+G(t-\Delta t)}{2\Delta t}, \quad G''(t)\approx \frac{G(t+\Delta t)+G(t-\Delta t)-2G(t)}{\Delta t^2}.
\end{equation}
For $D'\big(c(t)\big)$, we can use the Jacobian of the decoder as implemented by Pytorch or difference approximation that is similar to $G'(t)$.
To summarize this part, the overall loss function of our interpolation network is:
\begin{equation}
\label{7}
L_{total}=\lambda_{1}L_{conspeed}+ \lambda_{2}L_{geo}+\lambda_{3}L_{min},
\end{equation}
where $\lambda_{1}$, $\lambda_{2}$ and $\lambda_{3}$ are the weights to balance these three losses. They have a default setting as $\lambda_{1}=1$, $\lambda_{2}=0.001$ and $\lambda_{3}=10$ to ensure a robust performance. 
Under this loss constraint, we can generate interpolations moving along the minimizing geodesic with constant speed and thus fulfill a uniform interpolation. The overall geodesic learning algorithm is given in Algorithm~\ref{algorithm}.
\begin{algorithm}[H]
	\caption{\textbf{Geodesic Learning}} 
	\label{algorithm}	
	\begin{algorithmic}[1]
		\Require Two random points $x_0$, $x_1$ on dataset; $N$ points of time $t_i=\frac{i-1}{N-1}(1\leq i\leq N)$; The number of iterations of the interpolation network $n_{iter}$; The weight parameters $\lambda_{1}, \lambda_{2}, \lambda_{3}$; Delta-time $\Delta t$.
		\State Train an autoencoder by optimizing Eq.~(\ref{1});
		\State Obtain latent embeddings with encoder E of the trained autoencoder: $z_0=E(x_0)$, $z_1=E(x_1)$;
		\State Initialize interpolation network by setting $a=0, b=0, c=z_1-z_0, d=z_0$;
		\For{$j=1$ to $n_{iter}$}
		\State Obtain latent representations of interpolations $c(t_i)$ using interpolation network. \State Decode $c(t_i)$ using decoder D to get  $G(t_i)=D(c(t_i)) (1\leq i\leq N)$.
		\State Based on Eq.~(\ref{6}), compute $G'(t_i)$ and $G''(t_i)$ with $\Delta t$.
		\State Compute the overall loss $L_{total}$ using Eq.~(\ref{7}).
		\State Update interpolation network's parameters $a, b, c, d$ with RMSProp optimizer.
		\EndFor
		\State \Return Interpolations $G(t_i)$
	\end{algorithmic}
\end{algorithm}
\section{Experimental Results}
In this section, we present experiments on the geodesic generation and image interpolation to demonstrate the effectiveness of our method.

\subsection{Geodesic Generation}
First, we do experiments on 3-dimensional datasets since their geodesics can be better visualized. 
We choose the semi-sphere and swiss-roll as our data manifolds. 
\begin{figure*}[!th]
	\footnotesize
	\vspace{-5pt}
	\centering
	\renewcommand{\tabcolsep}{1pt} \renewcommand{\arraystretch}{0.1} \begin{tabular}{cccc}
		\includegraphics[width=0.25\linewidth]{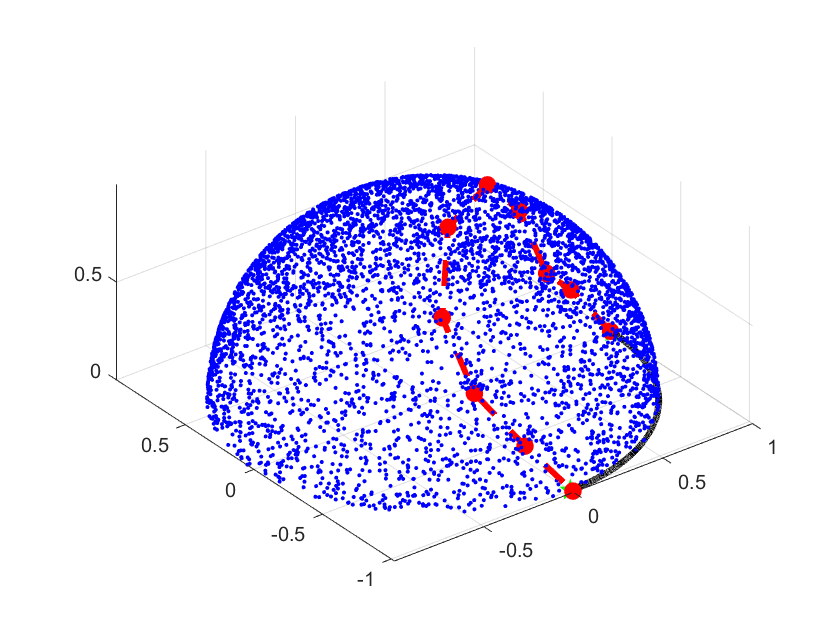} &
		\includegraphics[width=0.25\linewidth]{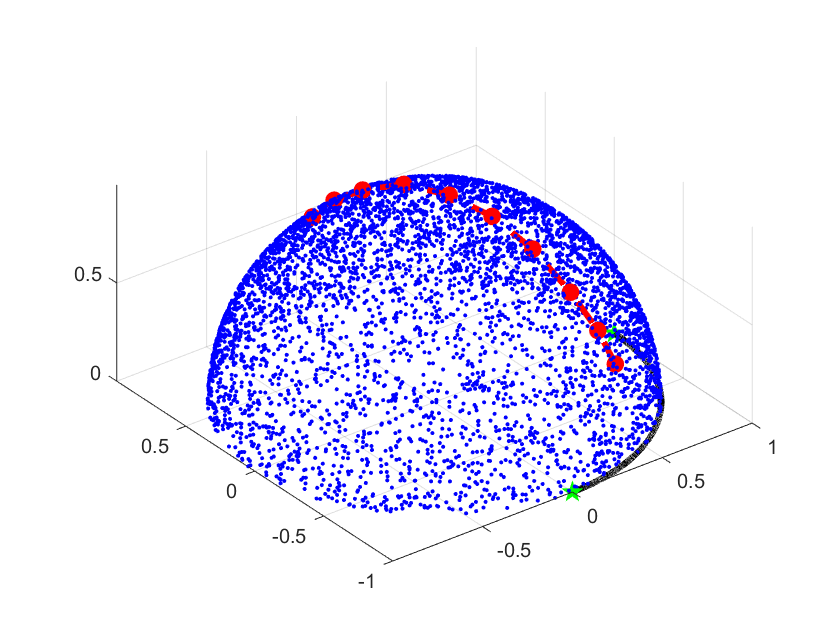} &
		\includegraphics[width=0.25\linewidth]{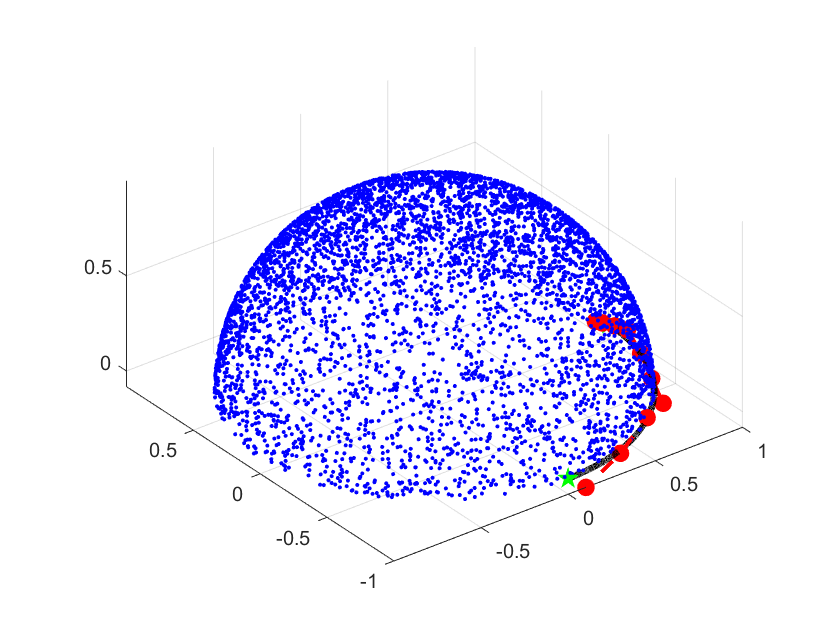}&
		\includegraphics[width=0.25\linewidth]{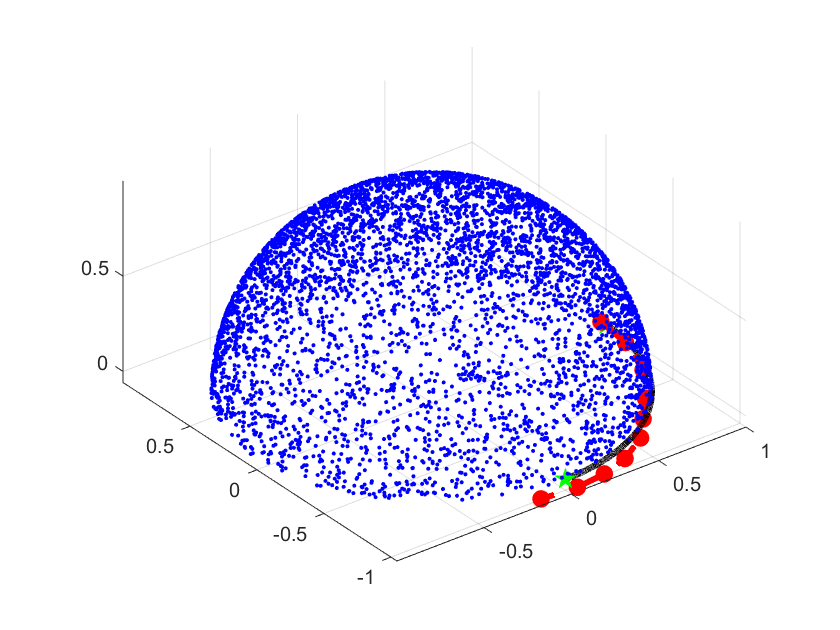}
		\\
		\vspace{-0cm} 
		\includegraphics[width=0.25\linewidth]{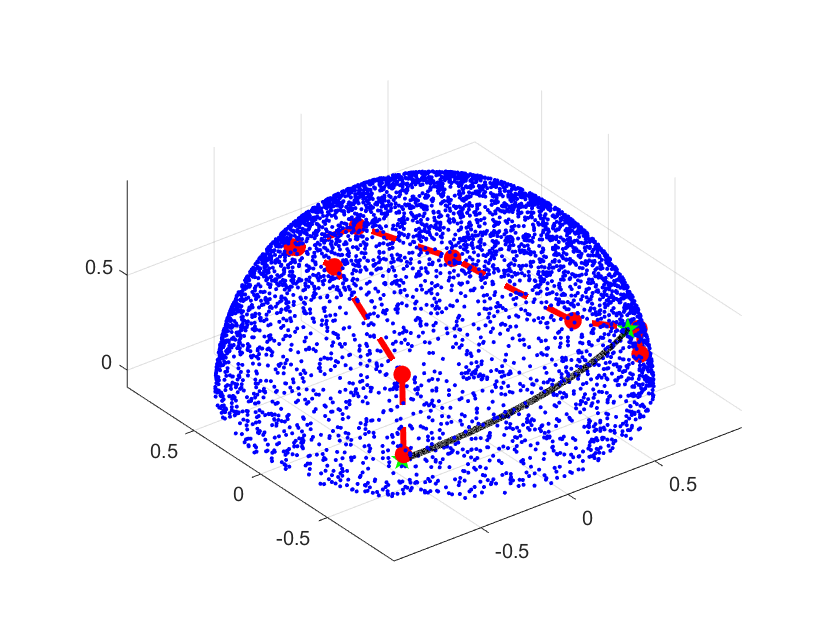} &
		\includegraphics[width=0.25\linewidth]{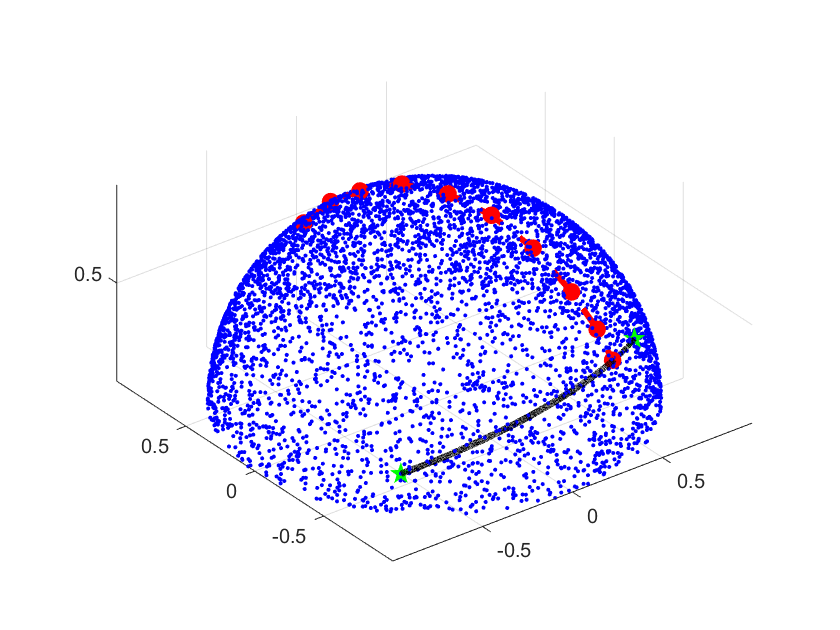}&
		\includegraphics[width=0.25\linewidth]{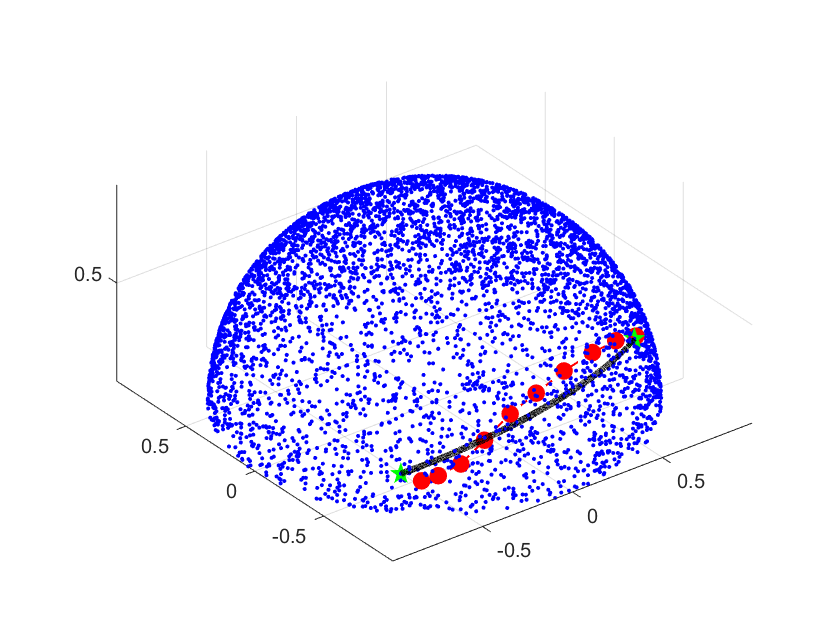}&
		\includegraphics[width=0.25\linewidth]{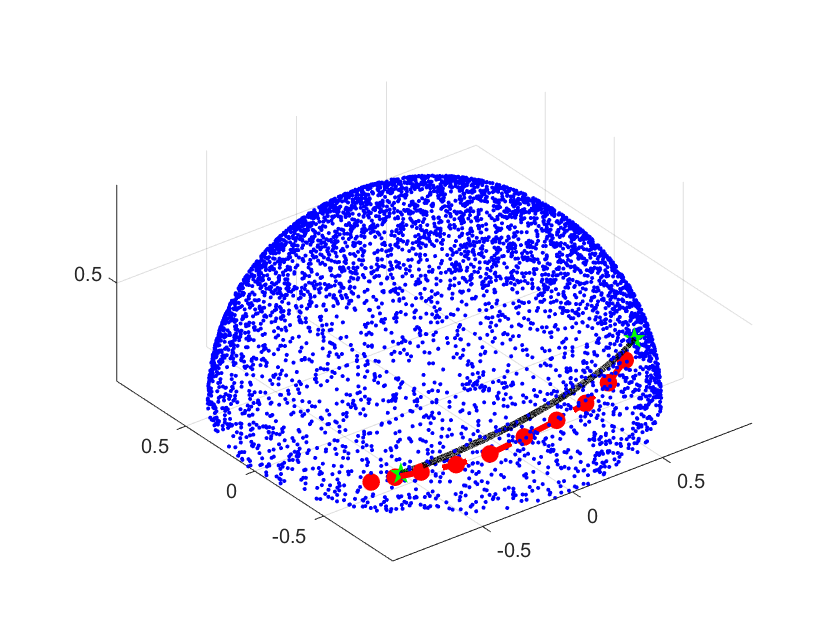}
		\\
		(a) AAE~\cite{makhzani2015adversarial} & (b) A\&H~\cite{arvanitidis2019fast} & (c) Chen's method~\cite{chen2018Metrics}& (d) ours \\
	\end{tabular}
	\caption{Results of different interpolation methods on semi-sphere Dataset.
	}
	\label{results_circle} \vspace{-10pt}
\end{figure*}

\begin{figure*}[!ht]
	\footnotesize
	\vspace{-5pt}
	\centering
	\renewcommand{\tabcolsep}{1pt} \renewcommand{\arraystretch}{0.1} \begin{tabular}{ccccc}
		\includegraphics[width=0.2\linewidth]{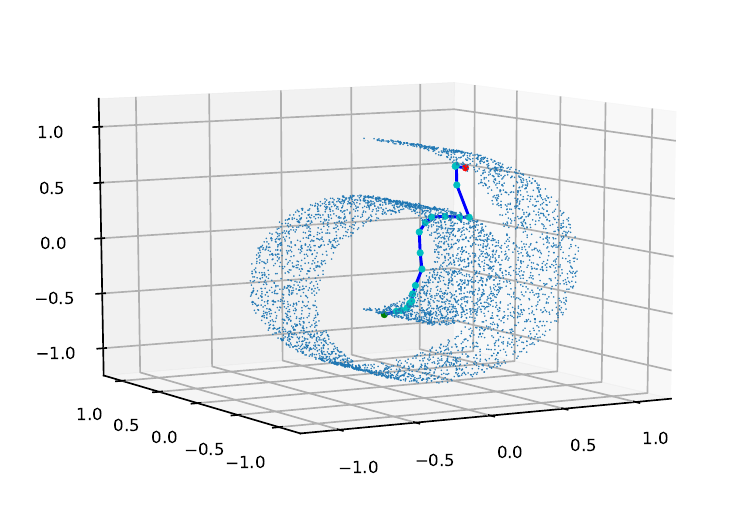} &
		\includegraphics[width=0.2\linewidth]{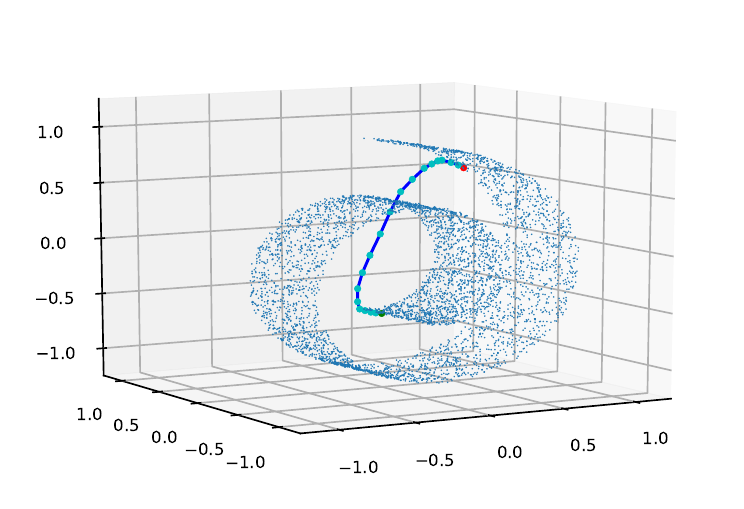} &
		\includegraphics[width=0.2\linewidth]{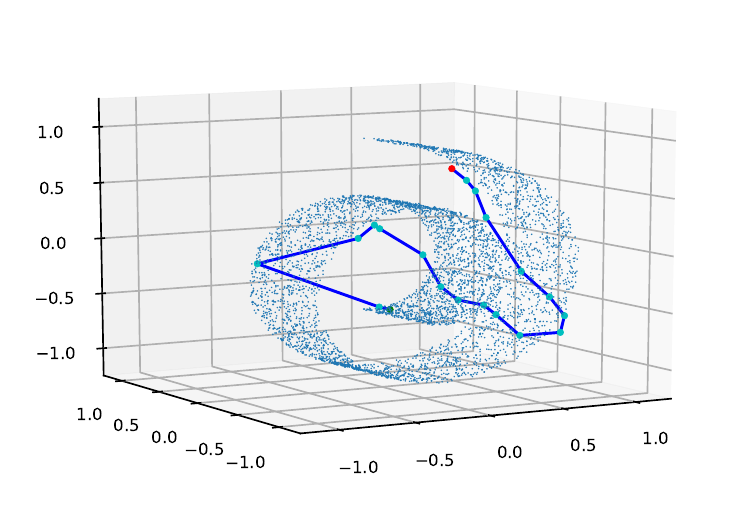}&
		\includegraphics[width=0.2\linewidth]{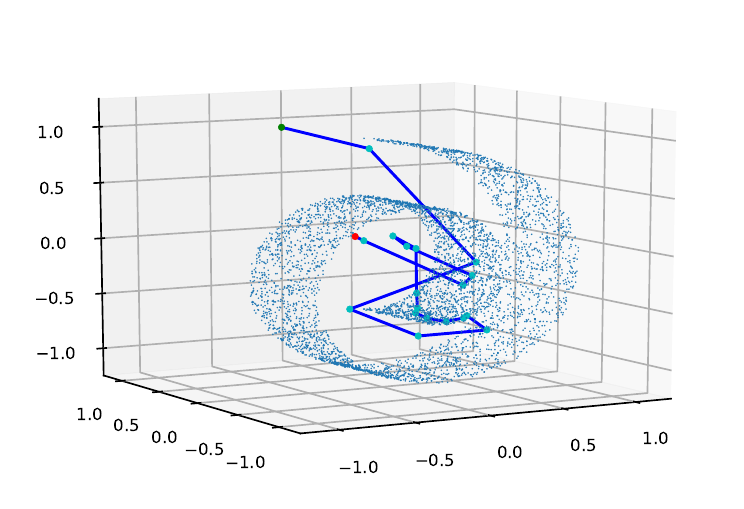}&
		\includegraphics[width=0.2\linewidth]{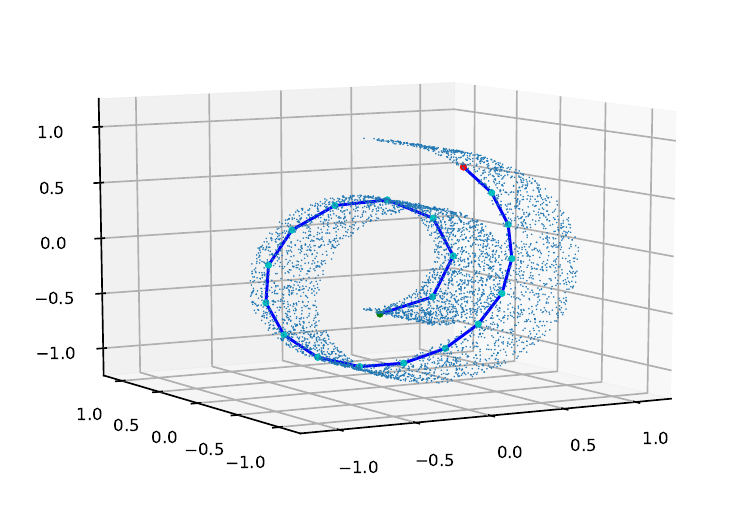}
		\\
		(a) AAE~\cite{makhzani2015adversarial} & (b) ACAI~\cite{berthelot2019Understandinga} & (c) GAIA~\cite{sainburg2018generative}& (d) Chen's method~\cite{chen2019homomorphic}& (e) ours \\
	\end{tabular}
	\vspace{-5pt}
	\caption{
		Results of different interpolation methods on swiss-roll Dataset.
	}
	\label{results_swissroll} \vspace{-5pt}
\end{figure*}
\noindent \textbf{Semi-sphere dataset.} We randomly sample 4,956 points subjecting to the uniform distribution on semi-sphere. In  Fig.~\ref{results_circle}, we compare our approach with other interpolation methods,~i.e., AAE~\cite{goodfellow2014generative}, A\&H~\cite{arvanitidis2019fast} and Chen's method~\cite{chen2018Metrics}. 
From Riemannian geometry, we know the geodesic of a sphere under our defined Riemannian metric is a circular arc or part of it. 
The center of this circular arc is the center of this sphere. 
AAE cannot guarantee that the curve connecting two points is a geodesic. A\&H can find the shortest path connecting two corresponding reconstructed endpoints.
But the endpoints are inconsistent with the original inputs due to the uncertainty of their VAE network and their stochastic Riemannian metric. Chen's method can generate interpolations along a geodesic, but they cannot fulfill a uniform interpolation. 
In Fig.~\ref{results_circle}, we observe that our method can generate uniform interpolation along a fairly accurate geodesic on semi-sphere based on the defined Riemannian metric. 
\begin{wraptable}{r}{0.4\linewidth}
	\vspace{-9pt}
	\centering
	\caption{The estimated distance of interpolation curve trained with different losses proposed in our method.}
	\label{table_ablation_study}
	\vspace{-5pt}
	\begin{tabular}{cc}
		\toprule
		Loss & \multicolumn{1}{l}{Distance} \\
		\midrule
		Linear & 1.3110 \\
		$L_{conspeed}$ & 1.4710 \\
		$L_{min}$ & 1.1571 \\
		$L_{conspeed}+L_{min}$ & 1.1277 \\
		$L_{conspeed}+L_{geo}$ & 6.0460 \\
		$L_{conspeed}+L_{geo}+L_{min}$ & 1.1028 \\
		real geodesic & \textbf{1.0776}\\
		\bottomrule
	\end{tabular} 
	\vspace{-10pt}
\end{wraptable}
\noindent  \textbf{Swiss-roll dataset.}
We choose swiss-roll to demonstrate the effectiveness of our method on manifolds that have large curvature variations. We randomly sample 5,000 points subjecting to the uniform distribution on the swiss-roll manifold. We compare our approach with AAE~\cite{makhzani2015adversarial}, ACAI~\cite{berthelot2019Understandinga}, GAIA~\cite{sainburg2018generative} and the method that applies GAN on feature space proposed by~\cite{chen2019homomorphic}.
Fig.~\ref{results_swissroll} shows the experimental results. 
We observe that except our approach, other methods fail to generate interpolations within the data manifold because they cannot encode efficient latent embeddings for linear interpolations.

\begin{figure}[htbp]
	\vspace{-10pt}
	\footnotesize
	\centering
	\renewcommand{\tabcolsep}{1pt} \renewcommand{\arraystretch}{0.9} \begin{tabular}{ccc}
		\includegraphics[width=0.3\linewidth]{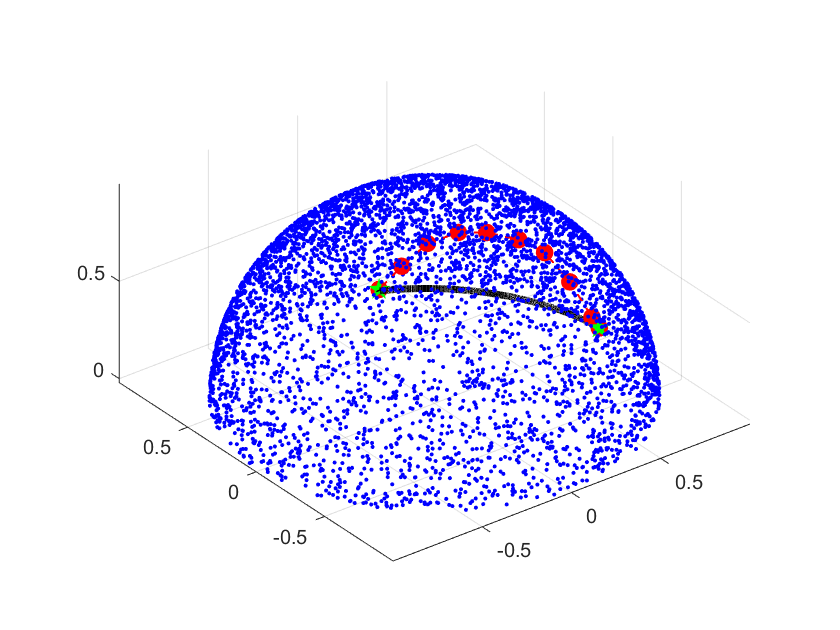} &
		\includegraphics[width=0.3\linewidth]{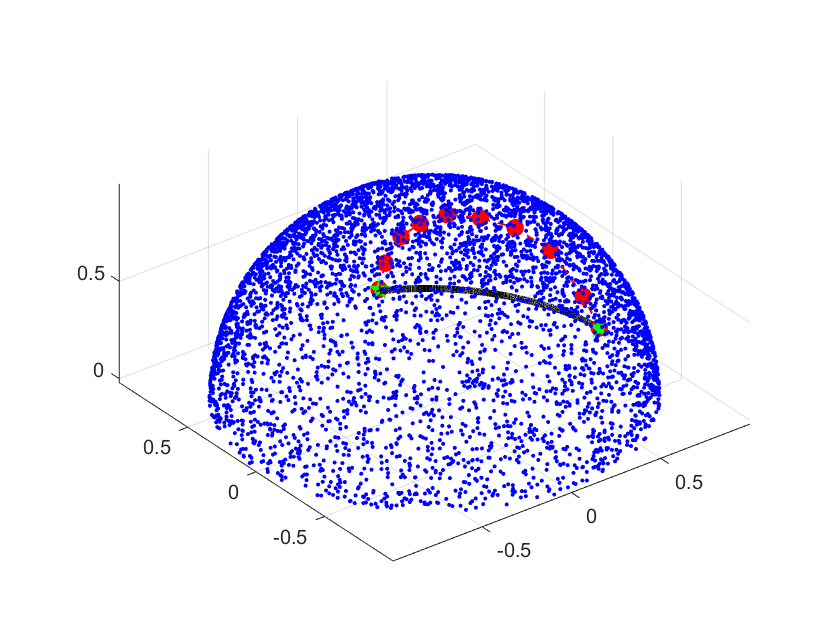} &
		\includegraphics[width=0.3\linewidth]{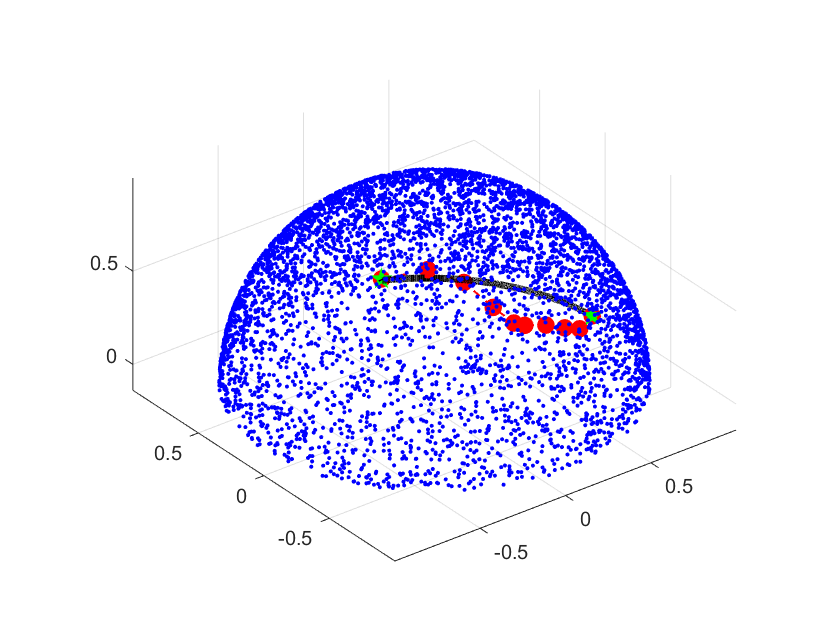}
		\\
		(a) linear interpolation & (b) $L_{conspeed}$ & (c) $L_{min}$\\
		\includegraphics[width=0.3\linewidth]{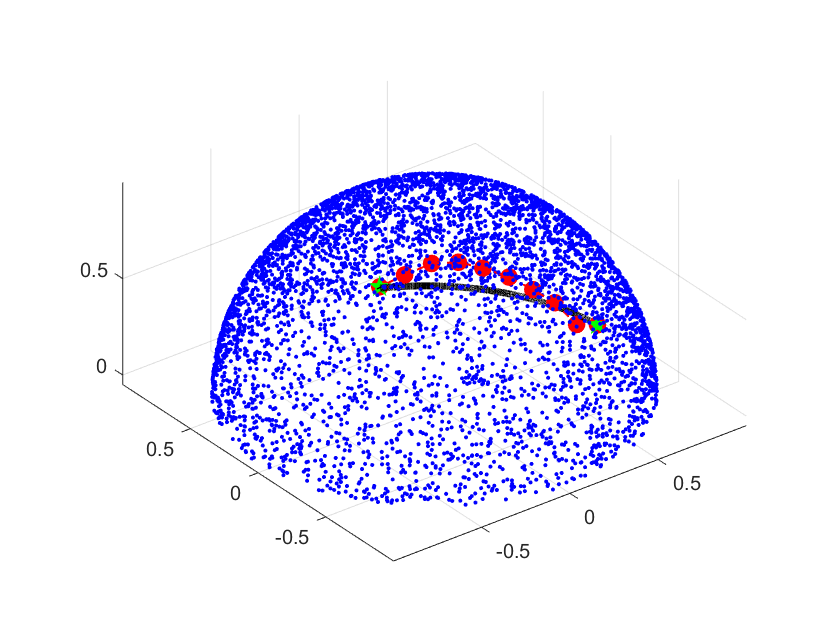} &
		\includegraphics[width=0.3\linewidth]{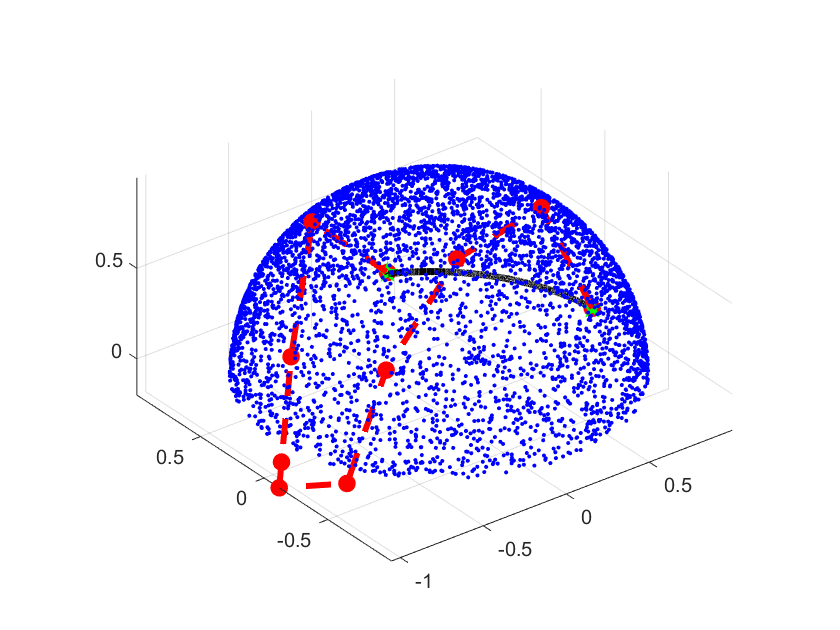} &
		\includegraphics[width=0.3\linewidth]{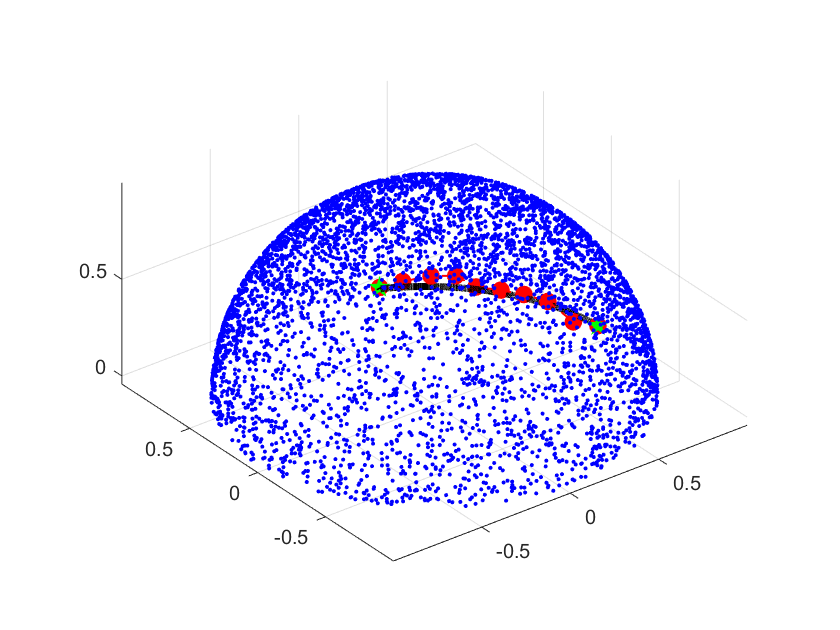}
		\\
	    (d) $L_{conspeed}+L_{min}$ &(e) $L_{conspeed}+L_{geo}$ & (f)  $L_{total}$\\
	\end{tabular}
	\vspace{-5pt}
	\caption{Ablation study of results trained with different losses proposed by our method.
	}
	\label{ablation_study} \vspace{-10pt}
\end{figure}

We do ablation studies on the semi-sphere dataset to investigate the effect of different losses proposed in our method, including the constant-speed loss, geodesic loss, and minimizing loss. Fig.~\ref{ablation_study} presents the results obtained by the combinations of different losses. We observe that the constant-speed loss can promote a uniform interpolation. Compared with linear interpolation, our network can generate a shorter path that is close to real geodesic without the geodesic loss. 
Although our network can quickly converge to a fairly satisfactory result, the accuracy of the geodesic is not optimal.
When incorporated with the geodesic loss, our interpolated points are fine-tuned to move along an accurate geodesic. 
The estimated curve length shown in Table~\ref{table_ablation_study} demonstrates our observation.
The generated curve trained with all three losses contributes the best result and the estimated curve length 1.1028 is the closest to the length of real geodesic, namely 1.0776.

\begin{figure*}[htbp]
		\begin{minipage}{0.25\linewidth}
			\begin{minipage}{\linewidth}
				\includegraphics[width=0.735\linewidth,height=0.6cm,right]{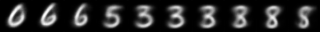}
			\end{minipage}
			\begin{minipage}{\linewidth}
				\includegraphics[width=0.735\linewidth,height=0.6cm,right]{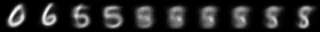}
			\end{minipage}
			\begin{minipage}{\linewidth}
				\centering
				\hspace{1cm}
				\includegraphics[width=0.745\linewidth,right]{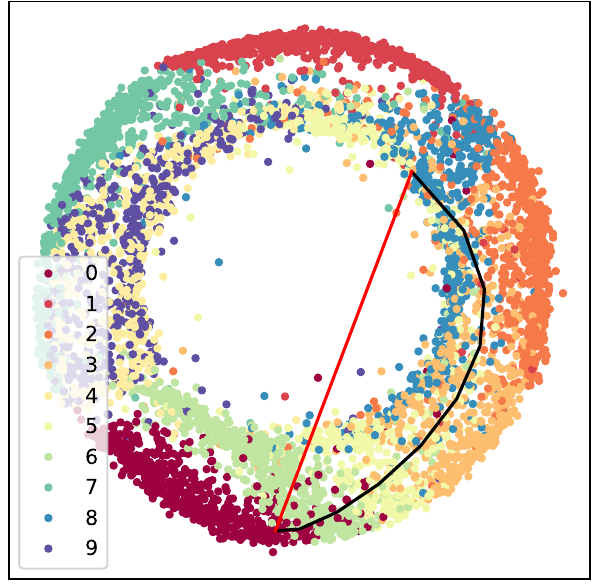}
			\end{minipage}
		\end{minipage}%
		\begin{minipage}{0.25\linewidth}
			\begin{minipage}{\linewidth}
				\includegraphics[width=0.735\linewidth,height=0.6cm,right]{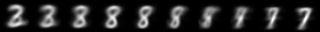}
			\end{minipage}
			\begin{minipage}{\linewidth}
				\includegraphics[width=0.735\linewidth,height=0.6cm,right]{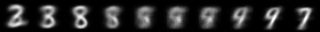}
			\end{minipage}
			\begin{minipage}{\linewidth}
				\centering
				\hspace{1cm}
				\includegraphics[width=0.745\linewidth,right]{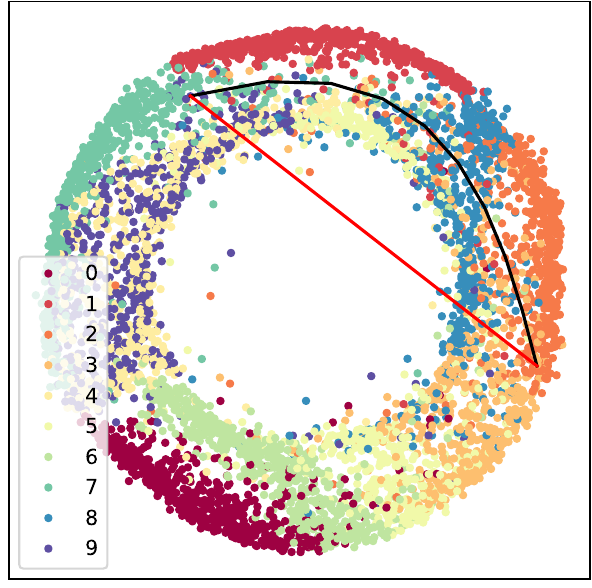}
			\end{minipage}
		\end{minipage}%
		\begin{minipage}{0.25\linewidth}
			\begin{minipage}{\linewidth}
				\includegraphics[width=0.735\linewidth,height=0.6cm,right]{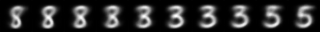}
			\end{minipage}
			\begin{minipage}{\linewidth}
				\includegraphics[width=0.735\linewidth,height=0.6cm,right]{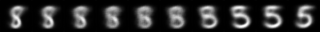}
			\end{minipage}
			\begin{minipage}{\linewidth}
				\centering
				\hspace{1cm}
				\includegraphics[width=0.745\linewidth,right]{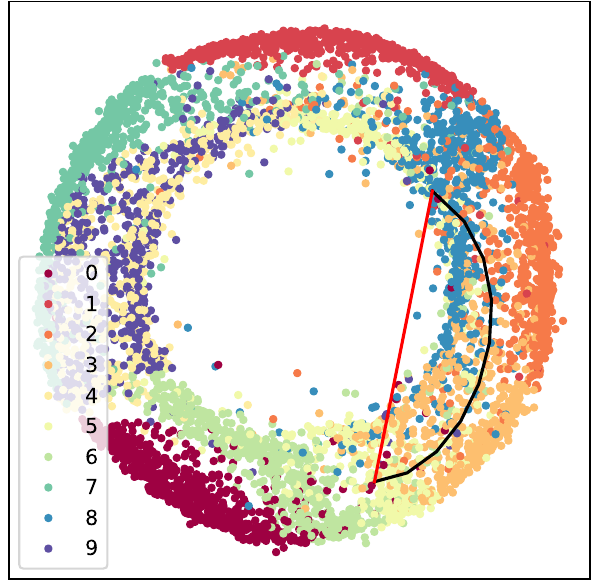}
			\end{minipage}
		\end{minipage}%
		\begin{minipage}{0.25\linewidth}
			\begin{minipage}{\linewidth}
				\includegraphics[width=0.735\linewidth,height=0.6cm,right]{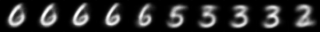}
			\end{minipage}
			\begin{minipage}{\linewidth}
				\includegraphics[width=0.735\linewidth,height=0.6cm,right]{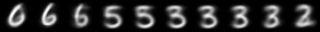}
			\end{minipage}
			\begin{minipage}{\linewidth}
				\centering
				\hspace{1cm}
				\includegraphics[width=0.745\linewidth,right]{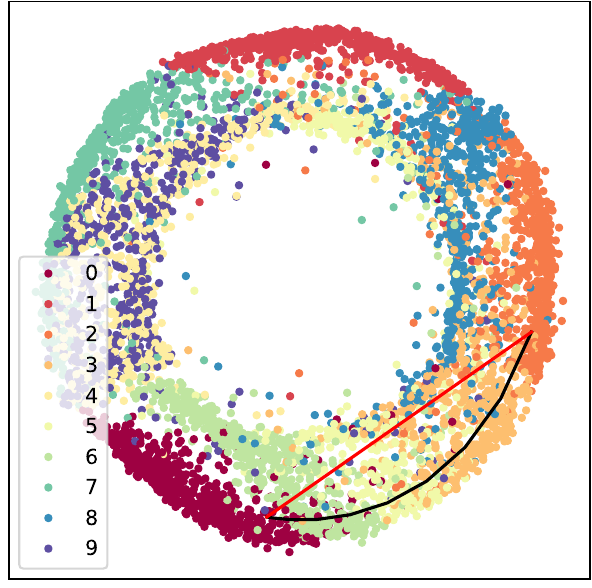}
			\end{minipage}
		\end{minipage}
		
	\caption{\textit{Top}: decoded images along linear (bottom) and geodesic (top) interpolants. \textit{Bottom}: the corresponding linear (red) and geodesic (black) interpolant trajectories on latent space.}
	\label{minist_scatter}\vspace{-10pt}
\end{figure*}

\begin{figure*}[htbp]
	\begin{subfigure}{0.5\textwidth}
		\begin{minipage}{0.5\linewidth}
			\begin{minipage}{\linewidth}
				\hspace{-0.15cm}
				\includegraphics[width=0.815\linewidth,height=0.6cm,right]{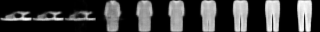}
			\end{minipage}
			\begin{minipage}{\linewidth}
				\hspace{-0.15cm}
				\includegraphics[width=0.815\linewidth,height=0.6cm,right]{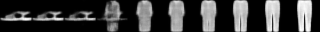}
			\end{minipage}
			\begin{minipage}{\linewidth}
				\centering
				\hspace{1cm}
				\includegraphics[width=\linewidth,height=1.8cm]{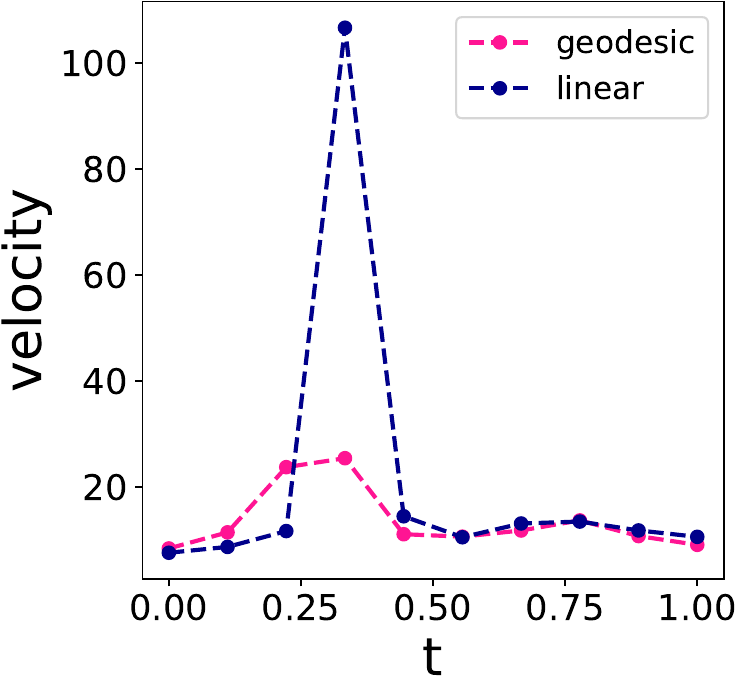}
			\end{minipage}
		\end{minipage}%
		\begin{minipage}{0.5\linewidth}
			\begin{minipage}{\linewidth}
				\hspace{-0.15cm}
				\includegraphics[width=0.815\linewidth,height=0.6cm,right]{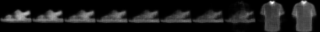}
			\end{minipage}
			\begin{minipage}{\linewidth}
				\hspace{-0.15cm}
				\includegraphics[width=0.815\linewidth,height=0.6cm,right]{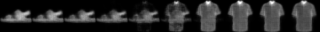}
			\end{minipage}
			\begin{minipage}{\linewidth}
				\centering
				\hspace{1cm}
				\includegraphics[width=\linewidth,height=1.8cm]{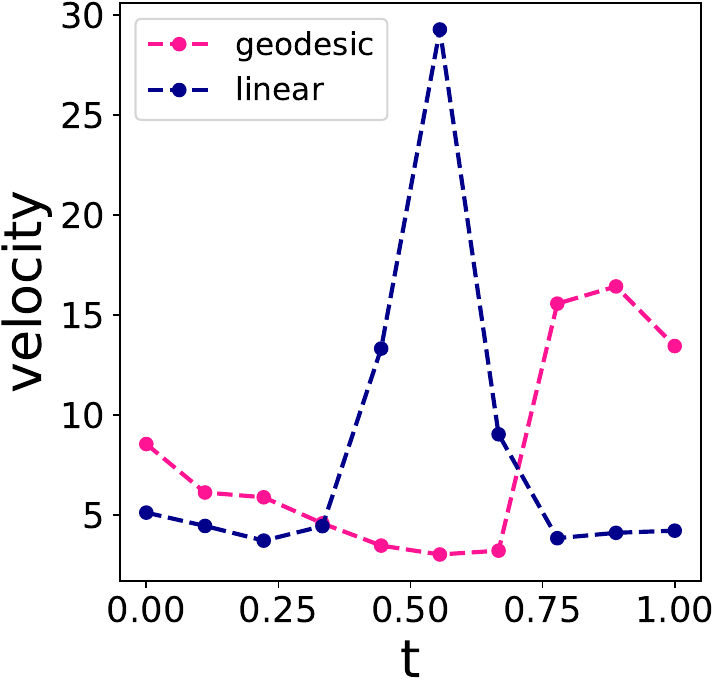}
			\end{minipage}
		\end{minipage}%
		
		\vspace{-0.2cm}
		\caption{VAE}
	\end{subfigure}%
	\hspace{0.15cm}
	\begin{subfigure}{0.5\textwidth}
		
		\begin{minipage}{0.5\linewidth}
			\begin{minipage}{\linewidth}
				\hspace{-0.15cm}
				\includegraphics[width=0.815\linewidth,height=0.6cm,right]{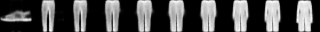}
			\end{minipage}
			\begin{minipage}{\linewidth}
				\hspace{-0.15cm}
				\includegraphics[width=0.815\linewidth,height=0.6cm,right]{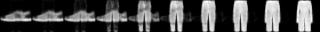}
			\end{minipage}
			\begin{minipage}{\linewidth}
				\centering
				\hspace{1cm}
				\includegraphics[width=\linewidth,height=1.8cm]{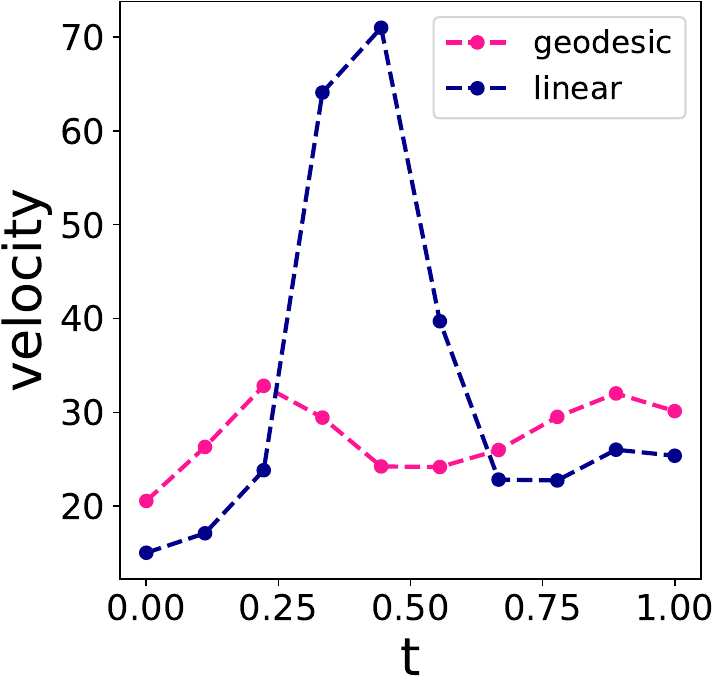}
			\end{minipage}
		\end{minipage}%
		\begin{minipage}{0.5\linewidth}
			\begin{minipage}{\linewidth}
				\hspace{-0.15cm}
				\includegraphics[width=0.815\linewidth,height=0.6cm,right]{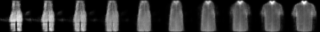}
			\end{minipage}
			\begin{minipage}{\linewidth}
				\hspace{-0.15cm}
				\includegraphics[width=0.815\linewidth,height=0.6cm,right]{mnist_velocity/image_aae_fashion/00500_90.png}
			\end{minipage}
			\begin{minipage}{\linewidth}
				\centering
				\hspace{1cm}
				\includegraphics[width=\linewidth,height=1.8cm]{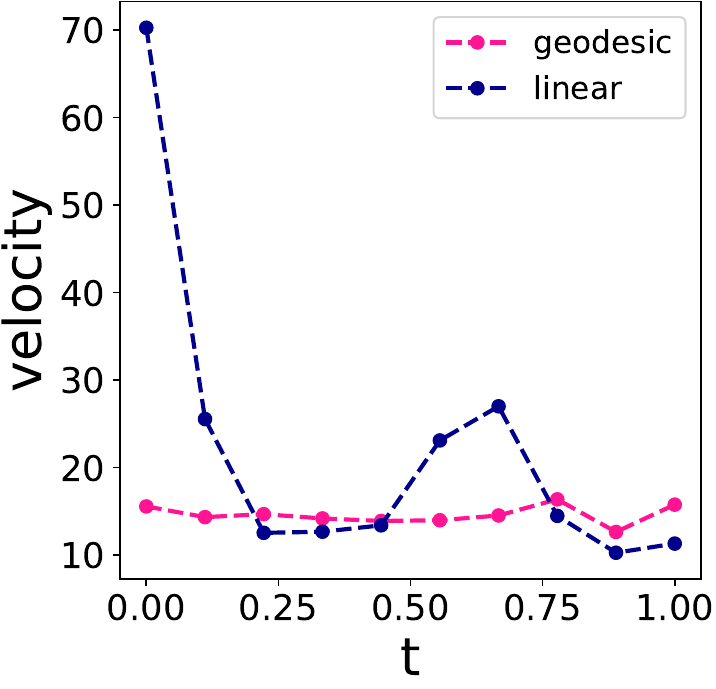}
			\end{minipage}
		\end{minipage}
		
		\vspace{-0.2cm}
		\caption{AAE}
	\end{subfigure}
	\caption{\textit{Top}: decoded images along linear (bottom) and geodesic (top) interpolants. \textit{Bottom}: the corresponding velocity curves of the interpolants.}
	\label{fashion_result}
\end{figure*}

\begin{figure*}[htbp]
	\footnotesize
	\centering
	\renewcommand{\tabcolsep}{1pt} \renewcommand{\arraystretch}{0.4} \begin{tabular}{cc}
		\includegraphics[width=0.45\linewidth]{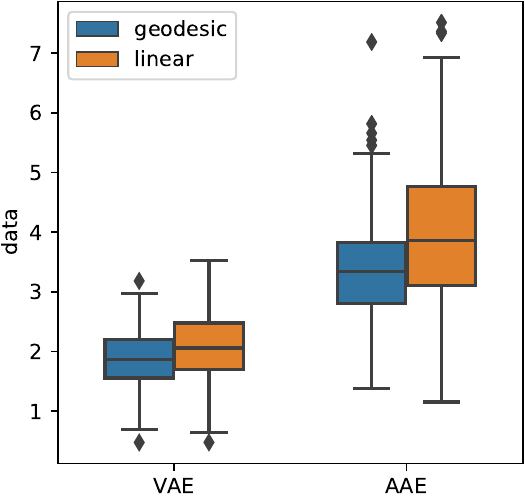} &
		\includegraphics[width=0.45\linewidth]{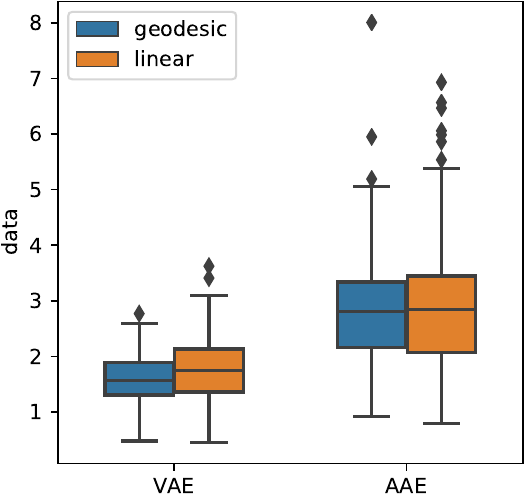}
		\\
		(a) MNIST&(b) Fashion-MNIST\\ 
	\end{tabular}
	\vspace{-5pt}
	\caption{The boxplots of the interpolant trajectory's length to linear and geodesic interpolations for both VAE and AAE.} 
	\label{boxplot} 
	\vspace{-5pt}
\end{figure*}

\subsection{Image interpolation}
To further demonstrate our model's effectiveness in image interpolation, we choose MNIST~\cite{lecun1998mnist} and Fashion-MNIST~\cite{xiao2017} datasets. They both consist of a training set of 60,000 examples and a test set of 10,000 examples associated with a label from 10 classes. We don't employ our manifold reconstruction method because the
concept of distance-based nearest neighbors is no longer meaningful when the
dimension goes sufficiently high~\cite{aggarwal2001surprising,beyer1999nearest}. Shao et al.~\cite{shaohang} argue that generated high-dimensional manifold has some curvature, but it
is close to being zero.
Therefore, we can directly employ variational autoencoder (VAE)~\cite{kingma2014AutoEncoding} and  adversarial autoencoder~(AAE)~\cite{makhzani2015adversarial} to obtain latent embeddings. For each dataset, we randomly select two images as the start point and endpoint in the data manifold. We employ different interpolation methods to realize an image interpolation between these two images. For the MNIST dataset, Fig.~\ref{minist_scatter} shows the generated geodesic and linear interpolant trajectories of an AAE where the imposed latent distribution is a 2-D donut.
We observe that our approach finds a trajectory in the latent space that is longer than a simple linear interpolation, but the intermediate points look more realistic
and provide a semantically meaningful morphing between its endpoints as it does not cross a region without data. Our interpolation method ensures our generated curves to follow the data manifold. From the fourth column in Fig.~\ref{minist_scatter}, we can see although the geodesic and linear trajectories are both on the data manifold, the interpolation results can also be very different. We provide more results on MNIST dataset in supplementary materials to verify the invariance of geodesic interpolation with different autoencoders.

For the Fashion-MNIST dataset, we also use the methods above to interpolate images between two selected images and show the results in Fig.~\ref{fashion_result}. We use the velocity defined in Section~\ref{Constant Speed Loss} to evaluate the smoothness of different interpolation methods. Our geodesic interpolation can obtain always recognizable images with a smoother speed while linear interpolation may generate some ghosting with two articles when transiting over different classes. This demonstrates our geodesic learning method can fulfill a uniform interpolation along a geodesic.

We further verify the characteristic of the geodesic's shortest path for our interpolation method. In Fig.~\ref{boxplot}, we present a comparison of the interpolant trajectory's length to different interpolation approaches for both VAE and AAE. We randomly choose 250 pairs of endpoints on data manifold for each valuation and approximate the trajectory's length using the summation of velocity at $t_i$ which is described in Section~\ref{Minimizing Geodesic Loss}. Fig.~\ref{boxplot} shows that our geodesic interpolation has smaller average length and variance on both MNIST and Fashion-MNIST datasets. This demonstrates our interpolation method can make the interpolation curve to possess more characteristics of geodesics compared with linear interpolation.

\section{Conclusion}
We explore the geometric structure of the data manifold by proposing a uniform interpolation constrained geodesic learning algorithm. We add prior geometric information to regularize our autoencoder to generate interpolants within the data manifold. We also propose a constant-speed loss and a minimizing geodesic loss to generate geodesic on the underlying manifold given two endpoints. Different from existing methods in which geodesic is defined as the shortest path on the graph connecting data points, our model defines geodesic consistent with the definition of a geodesic in Riemannian geometry. Experiments demonstrate our model can fulfill a uniform interpolation along the minimizing geodesics both on 3-D curved manifolds and high-dimensional image space.

\bibliography{example_paper}

\begin{thebibliography}{10}

\bibitem{aggarwal2001surprising}
Charu~C Aggarwal, Alexander Hinneburg, and Daniel~A Keim.
\newblock On the surprising behavior of distance metrics in high dimensional
  space.
\newblock In {\em International conference on database theory}, pages 420--434.
  Springer, 2001.

\bibitem{agustsson2019Optimal}
Eirikur Agustsson, Alexander Sage, Radu Timofte, and Luc Van~Gool.
\newblock Optimal transport maps for distribution preserving operations on
  latent spaces of {{Generative Models}}.
\newblock In {\em Proceedings of the 7th {{International Conference}} on
  {{Learning Representations}}({{ICLR}})}, 2019.

\bibitem{arvanitidis2018Latent}
Georgios Arvanitidis, Lars~Kai Hansen, and S{\o}ren Hauberg.
\newblock Latent {{Space Oddity}}: On the {{Curvature}} of {{Deep Generative
  Models}}.
\newblock In {\em Proceedings of the 6th {{International Conference}} on
  {{Learning Representations}}({{ICLR}})}, 2018.

\bibitem{arvanitidis2019fast}
Georgios Arvanitidis, S{\o}ren Hauberg, Philipp Hennig, and Michael Schober.
\newblock Fast and {{Robust Shortest Paths}} on {{Manifolds Learned}} from
  {{Data}}.
\newblock In {\em Proceedings of the 22nd {{International Conference}} on
  {{Artificial Intelligence}} and {{Statistics}} ({{AISTATS}})}, page~10, 2019.

\bibitem{berthelot2019Understandinga}
David Berthelot, Colin Raffel, Aurko Roy, and Ian Goodfellow.
\newblock Understanding and improving interpolation in autoencoders via an
  adversarial regularizer.
\newblock In {\em Proceedings of the 7th {{International Conference}} on
  {{Learning Representations}}({{ICLR}})}, 2019.

\bibitem{beyer1999nearest}
Kevin Beyer, Jonathan Goldstein, Raghu Ramakrishnan, and Uri Shaft.
\newblock When is “nearest neighbor” meaningful?
\newblock In {\em International conference on database theory}, pages 217--235.
  Springer, 1999.

\bibitem{bishop1997magnification}
Christopher~M Bishop, Markus Svens'~en, and Christopher~KI Williams.
\newblock Magnification factors for the som and gtm algorithms.
\newblock In {\em Proceedings 1997 Workshop on Self-Organizing Maps}, 1997.

\bibitem{carmo1992riemannian}
Manfredo Perdigao~do Carmo.
\newblock {\em Riemannian geometry}.
\newblock Birkh{\"a}user, 1992.

\bibitem{chen2019fast}
Nutan Chen, Francesco Ferroni, Alexej Klushyn, Alexandros Paraschos, Justin
  Bayer, and Patrick van~der Smagt.
\newblock Fast approximate geodesics for deep generative models.
\newblock In {\em International Conference on Artificial Neural Networks},
  2019.

\bibitem{chen2018Metrics}
Nutan Chen, Alexej Klushyn, Richard Kurle, Xueyan Jiang, Justin Bayer, and
  Patrick {van der Smagt}.
\newblock Metrics for {{Deep Generative Models}}.
\newblock {\em in Proceedings of the 21st International Conference on
  Artificial Intelligence and Statistics (AISTATS)}, 2018.

\bibitem{Introduction_to_Riemann_Geometry}
Weihuan Chen and Xingxiao Li.
\newblock {\em Introduction to Riemann Geometry: Volume I}.
\newblock Peking University Press, 2004.

\bibitem{chen2019homomorphic}
Ying-Cong Chen, Xiaogang Xu, Zhuotao Tian, and Jiaya Jia.
\newblock Homomorphic latent space interpolation for unpaired image-to-image
  translation.
\newblock In {\em Proceedings of the IEEE Conference on Computer Vision and
  Pattern Recognition}, pages 2408--2416, 2019.

\bibitem{goodfellow2014generative}
Ian Goodfellow, Jean Pouget-Abadie, Mehdi Mirza, Bing Xu, David Warde-Farley,
  Sherjil Ozair, Aaron Courville, and Yoshua Bengio.
\newblock Generative adversarial nets.
\newblock In {\em Advances in neural information processing systems}, pages
  2672--2680, 2014.

\bibitem{kingma2014AutoEncoding}
Diederik~P. Kingma and Max Welling.
\newblock Auto-{{Encoding Variational Bayes}}.
\newblock In {\em Proceedings of the 2nd {{International Conference}} on
  {{Learning Representations}} ({{ICLR}})}, 2014.

\bibitem{lecun1998mnist}
Yann LeCun.
\newblock The mnist database of handwritten digits.
\newblock {\em http://yann. lecun. com/exdb/mnist/}, 1998.

\bibitem{makhzani2015adversarial}
Alireza Makhzani, Jonathon Shlens, Navdeep Jaitly, Ian Goodfellow, and Brendan
  Frey.
\newblock Adversarial autoencoders.
\newblock {\em arXiv preprint arXiv:1511.05644}, 2015.

\bibitem{pai2019DIMAL}
Gautam Pai, Ronen Talmon, Alex Bronstein, and Ron Kimmel.
\newblock {{DIMAL}}: {{Deep Isometric Manifold Learning Using Sparse Geodesic
  Sampling}}.
\newblock {\em In 2019 IEEE Winter Conference on Applications of Computer
  Vision (WACV)}, 2019.

\bibitem{LLE}
Sam~T Roweis and Lawrence~K Saul.
\newblock Nonlinear dimensionality reduction by locally linear embedding.
\newblock {\em science}, 290(5500):2323--2326, 2000.

\bibitem{sainburg2018generative}
Tim Sainburg, Marvin Thielk, Brad Theilman, Benjamin Migliori, and Timothy
  Gentner.
\newblock Generative adversarial interpolative autoencoding: adversarial
  training on latent space interpolations encourage convex latent
  distributions.
\newblock {\em arXiv preprint arXiv:1807.06650}, 2018.

\bibitem{shaohang}
H.~{Shao}, A.~{Kumar}, and P.~T. {Fletcher}.
\newblock The riemannian geometry of deep generative models.
\newblock In {\em 2018 IEEE/CVF Conference on Computer Vision and Pattern
  Recognition Workshops (CVPRW)}, pages 428--4288, 2018.

\bibitem{isomap}
Joshua~B Tenenbaum, Vin De~Silva, and John~C Langford.
\newblock A global geometric framework for nonlinear dimensionality reduction.
\newblock {\em science}, 290(5500):2319--2323, 2000.

\bibitem{tolstikhin2018Wasserstein}
Ilya Tolstikhin, Olivier Bousquet, Sylvain Gelly, and Bernhard Schoelkopf.
\newblock Wasserstein {{Auto}}-{{Encoders}}.
\newblock {\em In Proceedings of the 6th International Conference on Learning
  Representations (ICLR)}, 2018.

\bibitem{xiao2017}
Han Xiao, Kashif Rasul, and Roland Vollgraf.
\newblock Fashion-mnist: a novel image dataset for benchmarking machine
  learning algorithms.
\newblock 2017.

\bibitem{yang2018Geodesic}
Tao Yang, Georgios Arvanitidis, Dongmei Fu, Xiaogang Li, and S{\o}ren Hauberg.
\newblock Geodesic {{Clustering}} in {{Deep Generative Models}}.
\newblock {\em arXiv:1809.04747 [cs, stat]}, September 2018.

\bibitem{ltsa}
Zhenyue Zhang and Hongyuan Zha.
\newblock Nonlinear dimension reduction via local tangent space alignment.
\newblock In {\em International Conference on Intelligent Data Engineering and
  Automated Learning}, pages 477--481. Springer, 2003.

\end{thebibliography}
\bibliographystyle{plain}

\newpage
\makeatletter
\newcommand{\nipstophline}{%
	\noalign {\ifnum 0=`}\fi \hrule height 4pt
	\futurelet \reserved@a \@xhline
}
\newcommand{\nipsbottomhline}{%
	\noalign {\ifnum 0=`}\fi \hrule height 1pt
	\futurelet \reserved@a \@xhline
}
\begin{table}
	\setlength{\tabcolsep}{0.2cm}
	\begin{tabular}{p{0.97\columnwidth}}
		\nipstophline
		\rule{0pt}{1.0cm}
		\centering
		\Large{\textbf{Supplementary Materials: Uniform Interpolation Constrained Geodesic Learning on Data Manifold}}
		\vspace{4pt}
	\end{tabular}
\end{table}
\begin{table}
	\begin{tabular}{c}
		\nipsbottomhline
		~~~~~~~~~~~~~~~~~~~~~~~~~~~~~~~~~~~~~~~~~~~~~~~~~~~~~~~~~~~~~~~~~~~~~~~~~~~~~~~~~~~~~~~~~~~~~~~~~~~~~~~~~~~~~~~~~~~~~~~~~~~~~~~~~~~~~~~~~~~~~~~~~~~~~~~~~~
	\end{tabular}
\end{table}
\appendix
\section{Theoretical  part}
\newtheorem{lem}{Lemma}
\newtheorem*{proof}{Proof}
\begin{lem}\label{1.1}
 If $\gamma: I \to M$ is a geodesic on Riemannian manifold $M$, then the length of the tangent vector $\frac{d\gamma}{dt}$ is constant.
\end{lem}
\begin{proof}
Suppose $D$ is the Riemannian connection of $M$, then we have
	\begin{equation}
	\frac{d}{dt}\Big\langle\frac{d\gamma}{dt},\frac{d\gamma}{dt}\Big\rangle=2\Big\langle\frac{D}{dt}\frac{d\gamma}{dt},\frac{d\gamma}{dt}\Big\rangle=0,
	\end{equation}
Thus,
\begin{equation}
|\gamma'(t)|=C~(constant)
\end{equation}
\end{proof}
Proof of Theorem~\ref{4.1}.

\begin{proof}
	Suppose the Riemannian metric on $\mathcal{X}$ is induced from $\mathbb{R}^N$ by identity mapping $i$. $[x_1,x_2,\cdots,x_N]$ is the Cartesian coordinate system of $\mathbb{R}^N$. $\big[x_1(t),x_2(t),\cdots,x_N(t)\big]$ is the Cartesian coordinate of $\gamma(t)$. Based on the characteristic of $di$, we obtain the corresponding tangent vector of $\gamma'(t)$ in $T_{\gamma(t)}\mathbb{R}^N$:
\begin{equation}
\label{10}
di_{\gamma(t)}\big(\gamma'(t)\big)= \frac{dx_i(t)}{dt}\frac{\partial}{\partial x^i},\quad\footnote{Here we adopt the Einstein summation convention that repeated indices imply summation. This convention is adopted in the sequel. }
\end{equation}
According to the canonical metric on $\mathbb{R}^N$, we have 
\begin{equation}
\Big\vert di_{\gamma(t)}\big(\gamma'(t)\big)\Big\vert = \sqrt{\Big \langle \frac{dx_i(t)}{dt}\frac{\partial}{\partial x^i},\frac{dx_i(t)}{dt}\frac{\partial}{\partial x^i}\Big\rangle}=\sqrt{\sum_{i=1}^{N}\Big(\frac{dx_i(t)}{dt}\Big)^2}.
\end{equation}
Because $\gamma(t)$ is a geodesic on $\mathcal{X}$, from Lemma~\ref{1.1}, we know $|\gamma'(t)|$ is a constant. Considering identity mapping $i$ is a isometric immersion, we have 
\begin{equation}
\Big\vert di_{\gamma(t)}\big(\gamma'(t)\big)\Big\vert=|\gamma'(t)|.
\end{equation}
Thus,  $\sqrt{\sum_{i}\big(\frac{dx_i(t)}{dt}\big)^2}$ is a constant, for$~\forall t\in I$.
\end{proof}
Proof of Theorem~\ref{4.2}.
\begin{proof}
Suppose the Riemannian metric on $\mathcal{X}$ is induced from $\mathbb{R}^N$ by identity mapping $i$. $[x_1,x_2,\cdots,x_N]$ is the Cartesian coordinate system of $\mathbb{R}^N$. From the definition of $h$, we can get $h(z_1,z_2,\cdots,z_m)=x_1,x_2,\cdots,x_N$. Suppose $\frac{\partial h^j}{\partial z^i}$ is the $j$-th component of $\frac{\partial h}{\partial z^i}$, according to the characteristic of di, we can deduce the corresponding tangent vector of $\frac{\partial}{\partial z_i}$ on $T_{\gamma(t)}\mathbb{R}^N$ as follows:
\begin{equation}
di_{\gamma(t)}\Big(\frac{\partial}{\partial z^i}\Big)= \frac{\partial h^j}{\partial z^i} \frac{\partial}{\partial x^j}\Big|_{\gamma(t)},
\end{equation}
$\{di_{\gamma(t)}\big(\frac{\partial}{\partial z^i}\big), 1\leqslant i\leqslant N\}$ spans a tangent space $T_{\gamma(t)}i(\mathcal{X})$ at $\gamma(t)$. Suppose $D$ is the Riemannian connection on $\mathcal{X}$. According to the definition of $D\big(f_*(\gamma'(t))\big)$ (refer to Chapter 2, Eq.(3.1), in~\cite{Introduction_to_Riemann_Geometry}), we have
\begin{equation}
\frac{dD\big(di_{\gamma(t)}\big(\gamma'(t)\big)\big)}{dt}=D_{di_{\gamma(t)}(\gamma'(t))}di_{\gamma(t)}(\gamma'(t))=\Big(\frac{d\big(di_{\gamma(t)}(\gamma'(t))\big)}{dt}\Big)^{\top}.
\end{equation}
The first equality holds by Remark 3.2 of Chapter 2 in~\cite{Introduction_to_Riemann_Geometry}. $\top$ denotes the orthographic projection from $\mathbb{R}^N$ to the tangent space on $i(\mathcal{X})$. Because Riemannian connection is invariant under isometric condition (see Theorem 4.6 of Chapter 2 in~\cite{Introduction_to_Riemann_Geometry}), we have 
\begin{equation}
di_{\gamma(t)}(D_{\gamma'(t)}\gamma'(t))=D_{di_{\gamma(t)}(\gamma'(t))}di_{\gamma(t)}(\gamma'(t))=\Big(\frac{d\big(di_{\gamma(t)}(\gamma'(t))\big)}{dt}\Big)^{\top}.
\end{equation}
Combined with Eq.~(\ref{10}), we can deduce 
\begin{equation}
\Big(\frac{d\big(di_{\gamma(t)}(\gamma'(t))\big)}{dt}\Big)^{\top}=\Big(\frac{d^2x_i(t)}{dt^2}\frac{\partial}{\partial x^i}\Big)^{\top}.
\end{equation}
Therefore, based on the definition of geodesic, $\gamma: I\to \mathcal{X}$ is a geodesic on $\mathcal{X}$, if and only if:
\begin{equation}
\Big(\frac{d^2x_i(t)}{dt^2}\frac{\partial}{\partial x^i}\Big)^{\top}=0.
\end{equation}
It means $\frac{d^2x_i(t)}{dt^2}\frac{\partial}{\partial x^i}$ is orthogonal to tangent space $T_{\gamma(t)}i(\mathcal{X})$. That is, 
\begin{equation}
\label{19}
\Big\langle\frac{d^2x_i(t)}{dt^2}\frac{\partial}{\partial x^i}, di_{\gamma(t)}\Big(\frac{\partial}{\partial z^i}\Big)\Big\rangle=0,\quad 1\leqslant i\leqslant m
\end{equation}
Eq.~(\ref{19}) is eqivalent to 
\begin{equation}
\sum_{j=1}^{N}\frac{d^2x_j(t)}{dt^2}\cdot\frac{\partial h^j}{\partial z^i}=0,\quad 1\leqslant i\leqslant m
\end{equation}
We write it in the form of matrix multiplication,
\begin{equation}
[x_1''(t),x_2''(t),\cdots,x_N''(t)]\cdot \Big[\frac{\partial h}{\partial z^1}\Big|_{\gamma(t)},\frac{\partial h}{\partial z^2}\Big|_{\gamma(t)},\cdots,\frac{\partial h}{\partial z^m}\Big|_{\gamma(t)}\Big]=\bm{0}.
\end{equation}
\end{proof}
\newtheorem{them2}{Theorem}
\begin{them2}
\label{A1}
If a piece-wise differentiable curve $\gamma:[a,b] \to M$ \cite{carmo1992riemannian}, with parameter proportional to arc length, has length less or equal to the length of any other piece-wise differentiable curve joining $\gamma(a)$ to $\gamma(b)$ then $\gamma$ is a geodesic.
\end{them2}
Proof of Theorem~\ref{A1} can be referred to in corollary 3.9 of Chapter 3 in~\cite{carmo1992riemannian}
\section{Experimental part}
\subsection{Model Architecture}
For 3-dimensional datasets, we trained our interpolation network for 3000 iterations. Table~\ref{network1} describes the autoencoder architecture.

For MNIST and Fashion-MNIST datasets, we trained our interpolation for 30,000 iterations. Table~\ref{network2} and Table~\ref{network3} describe the network architecture of VAE and AAE respectively.
\begin{table}[]
	\centering
	\caption{The network architecture trained for 3-D datasets}
	\label{network1}
		\begin{tabular}{lcc}
			\hline
			\hline
			\multicolumn{3}{c}{Encoder}                                                              \\ \hline
			\hline
			Operation                       & \multicolumn{1}{l}{Input} & \multicolumn{1}{l}{Output} \\ \hline
			\multicolumn{1}{c}{FC}          & 3                         & 64                         \\
			\multicolumn{1}{c}{BN,PReLU,FC} & 64                        & 64                         \\
			\multicolumn{1}{c}{BN,PReLU,FC} & 64                        & 64                         \\
			BN,PReLU,FC                     & 64                        & 64                         \\
			BN,PReLU,FC                     & 64                        & 3                          \\
			\multicolumn{1}{c}{FC}          & 3                         & 2                          \\ \hline
			\hline
			\multicolumn{3}{c}{Decoder}                                                              \\ \hline
			\hline
			\multicolumn{1}{c}{BN,FC}       & 2                         & 64                         \\
			BN,ReLU,FC                      & 64                        & 64                         \\
			BN,ReLU,FC                      & 64                        & 64                         \\
			BN,ReLU,FC                      & 64                        & 64                         \\
			BN,ReLU,FC                      & 64                        & 64                         \\
			BN,ReLU,FC                      & 64                        & 3                          \\ \hline
			\hline
		\end{tabular}
	\end{table}
	\begin{table}[]
		\centering
		\caption{The network architecture of VAE trained for MNIST and Fashion-MNIST datasets}
		\label{network2}
		\begin{tabular}{lcccc}
			\hline \hline
			\multicolumn{5}{c}{Encoder}                                                                                                                                                               \\ \hline \hline
			\multicolumn{1}{l|}{}                       & Operation                                     & \multicolumn{1}{l}{Kernel} & \multicolumn{1}{l}{Strides} & \multicolumn{1}{l}{Feature maps} \\ \hline
			\multicolumn{1}{c|}{\multirow{7}{*}{$\mu$}}    & Conv2D,LReLU                                  & 4                          & 2                           & 64                               \\
			\multicolumn{1}{c|}{}                       & Conv2D,BN,LReLU                               & 4                          & 2                           & 128                              \\
			\multicolumn{1}{c|}{}                       & Conv2D,BN,LReLU                               & 4                          & 2                           & 256                              \\
			\multicolumn{1}{c|}{}                       & Conv2D,BN,LReLU                               & 4                          & 2                           & 512                              \\
			\multicolumn{1}{c|}{}                       & Reshape,Linear                                & \multicolumn{1}{l}{}       & \multicolumn{1}{l}{}        & 64                               \\
			\multicolumn{1}{c|}{}                       & BN,ReLU                                       & \multicolumn{1}{l}{}       & \multicolumn{1}{l}{}        & 64                               \\
			\multicolumn{1}{c|}{}                       & Reshape,Linear                                & \multicolumn{1}{l}{}       & \multicolumn{1}{l}{}        & 2                                \\ \hline
			\multicolumn{1}{l|}{\multirow{7}{*}{$\sigma$}} & Conv2D,LReLU                                  & 4                          & 2                           & 64                               \\
			\multicolumn{1}{l|}{}                       & Conv2D,BN,LReLU                               & 4                          & 2                           & 128                              \\
			\multicolumn{1}{l|}{}                       & Conv2D,BN,LReLU                               & 4                          & 2                           & 256                              \\
			\multicolumn{1}{l|}{}                       & Conv2D,BN,LReLU                               & 4                          & 2                           & 512                              \\
			\multicolumn{1}{l|}{}                       & Reshape,Linear                                & \multicolumn{1}{l}{}       & \multicolumn{1}{l}{}        & 64                               \\
			\multicolumn{1}{l|}{}                       & BN,ReLU                                       & \multicolumn{1}{l}{}       & \multicolumn{1}{l}{}        & 64                               \\
			\multicolumn{1}{l|}{}                       & Reshape,Linear                                & \multicolumn{1}{l}{}       & \multicolumn{1}{l}{}        & 2                                \\ \hline \hline
			\multicolumn{5}{c}{Decoder}                                                                                                                                                               \\ \hline \hline
			& \multicolumn{1}{l}{LReLU,ConvTranspose2D,BN}  & 4                          & 2                           & 512                              \\
			& \multicolumn{1}{l}{ReLU,ConvTranspose2D,BN}   & 4                          & 2                           & 256                              \\
			& \multicolumn{1}{l}{ReLU,ConvTranspose2D,BN}   & 4                          & 2                           & 128                              \\
			& \multicolumn{1}{l}{ReLU,ConvTranspose2D,BN}   & 4                          & 2                           & 64                               \\
			& \multicolumn{1}{l}{ReLU,ConvTranspose2D,Tanh} & 4                          & 2                           & 1                                \\ \hline \hline
		\end{tabular}
	\end{table}
\begin{table}[]
	\centering
	\caption{The network architecture of AAE trained for MNIST and Fashion-MNIST datasets}
	\label{network3}
	\begin{tabular}{cccc}
		\hline \hline
		\multicolumn{4}{c}{Encoder}                                                                                                                 \\ \hline \hline
		Operation                                     & \multicolumn{1}{l}{Kernel} & \multicolumn{1}{l}{Strides} & \multicolumn{1}{l}{Feature maps} \\ \hline
		Conv2D,LReLU                                  & 4                          & 2                           & 64                               \\
		Conv2D,BN,LReLU                               & 4                          & 2                           & 128                              \\
		Conv2D,BN,LReLU                               & 4                          & 2                           & 256                              \\
		Conv2D,BN,LReLU                               & 4                          & 2                           & 512                              \\
		Conv2D,BN                                     & 4                          & 2                           & 2                                \\ \hline \hline
		\multicolumn{4}{c}{Decoder}                                                                                                                 \\ \hline \hline
		\multicolumn{1}{l}{LReLU,ConvTranspose2D,BN}  & 4                          & 2                           & 512                              \\
		\multicolumn{1}{l}{ReLU,ConvTranspose2D,BN}   & 4                          & 2                           & 256                              \\
		\multicolumn{1}{l}{ReLU,ConvTranspose2D,BN}   & 4                          & 2                           & 128                              \\
		\multicolumn{1}{l}{ReLU,ConvTranspose2D,BN}   & 4                          & 2                           & 64                               \\
		\multicolumn{1}{l}{ReLU,ConvTranspose2D,Tanh} & 4                          & 2                           & 1                                \\ \hline \hline
	\end{tabular}
\end{table}
\subsection{More experimental results on MNIST dataset}
\begin{figure*}[htbp]
	\footnotesize
	\centering
	\renewcommand{\tabcolsep}{1pt} \renewcommand{\arraystretch}{0.4} \begin{tabular}{cc}
		\includegraphics[width=0.5\linewidth]{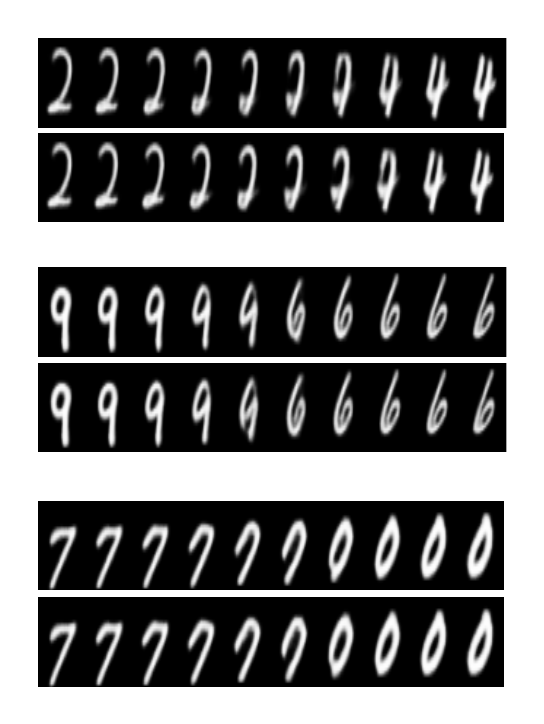}&
		\includegraphics[width=0.5\linewidth]{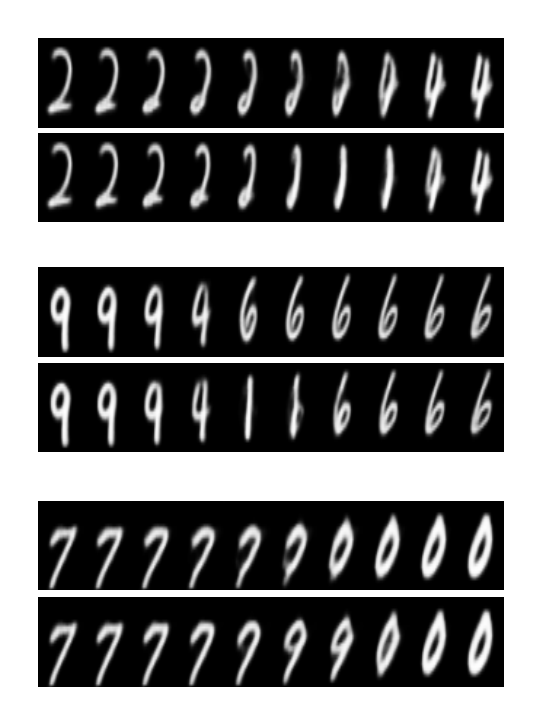}
		\\	
		(a) VAE &(b) AAE\\
	\end{tabular}
	\vspace{-5pt}
	\caption{\textbf{Visual interpolation results with different interpolation methods}:  interpolated images along linear (bottom) and geodesic (top) interpolants.}
	\label{MNIST_visual} \vspace{-10pt}
\end{figure*}

In this section, we provide more results for image interpolation on the MNIST dataset to verify the invariance of geodesic interpolation. Fig.~\ref{MNIST_visual} illustrates the visual interpolation results for both VAE and AAE being the autoencoder respectively. It worth mentioning that geodesics do not depend on the distributions of latent space because they only depend on the geometry structure of data manifold. Thus we should theoretically get the same interpolation results for different autoencoders if the interpolations are along geodesics. From Fig.~\ref{MNIST_visual}, we can observe for linear interpolation, different autoencoders result in very different results. For geodesic interpolation, although there is a slight difference between results with different autoencoders, the general transition tendency is consistent. Specifically, when we transit digit '2' to '4' shown in Fig.~\ref{MNIST_visual}, the linear interpolation with AAE will transit '2' to '1', then to '4', which is different from that with VAE transiting '2' to '4' directly. But geodesic interpolation with VAE and AAE both transit '2' to '4' directly and have a semantic transfer at similar sampling time. Other digit transitions have similar situations. These experimental results prove the invariance of geodesic interpolation with different autoencoders.
\end{document}